\pdfoutput=1

\documentclass[11pt]{article}

\usepackage{acl}

\usepackage{times}
\usepackage{latexsym}

\usepackage[T1]{fontenc}

\usepackage[utf8]{inputenc}

\usepackage{microtype}

\usepackage{inconsolata}

\usepackage{bm}
\usepackage{array}
\usepackage{float}
\usepackage{dsfont}
\usepackage{pifont}
\usepackage{amsmath}
\usepackage{amssymb}
\usepackage{amsfonts}
\usepackage{graphicx}
\usepackage{enumitem}
\usepackage{booktabs}
\usepackage{multirow}
\usepackage{multicol}
\usepackage{fontawesome5}
\usepackage[most]{tcolorbox}

\usepackage{tabularx}
\usepackage[figuresright]{rotating}
\usepackage{tablefootnote}

\DeclareSymbolFont{extraup}{U}{zavm}{m}{n}
\DeclareMathSymbol{\varheart}{\mathalpha}{extraup}{86}
\DeclareMathSymbol{\vardiamond}{\mathalpha}{extraup}{87}

\usepackage[vlined]{algorithm2e}

\definecolor{yellow}{HTML}{F6BD60}
\definecolor{white}{HTML}{F7EDE2}
\definecolor{pink}{HTML}{F5CAC3}
\definecolor{tale}{HTML}{84A59D}
\definecolor{red}{HTML}{F28482}
\definecolor{green}{HTML}{8EC07C}
\definecolor{orange}{HTML}{FE8019}
\definecolor{grey}{HTML}{EBDBB2}
\definecolor{brain}{HTML}{FFABBE}
\definecolor{blue}{HTML}{076678}
\definecolor{narrative}{HTML}{458588}

\newcolumntype{P}[1]{>{\centering\arraybackslash}p{#1}}

\newcommand{\ours}{{\fontfamily{qag}\selectfont{\textit{OpenToM}}}}
\newcommand{\tableinc}[1]{\colorbox{green}{#1}}
\newcommand{\tabledec}[1]{\colorbox{orange}{#1}}
\newcommand{\tablesame}[1]{\colorbox{grey}{#1}}

\title{\faPaw~\ours: A Comprehensive Benchmark for Evaluating Theory-of-Mind Reasoning Capabilities of Large Language Models}

\author{
    Hainiu Xu$^{1}$ \quad 
    Runcong Zhao$^{1}$ \quad 
    Lixing Zhu$^{1}$ \\
    \textbf{Jinhua Du}$^{2}$ \quad
    \textbf{Yulan He}$^{1,3}$ \\
    $^1$King's College London \quad\quad 
    $^2$Huawei London Research Centre \\
    $^3$The Alan Turing Institute \\
    {\tt \{hainiu.xu, runcong.zhao, lixing.zhu, yulan.he\}@kcl.ac.uk} \\
    {\tt \{jinhua.du\}@huawei.com}
}

\begin{document}
\maketitle
\begin{abstract}
Neural Theory-of-Mind (N-ToM), machine's ability to understand and keep track of the mental states of others, is pivotal in developing socially intelligent agents. However, prevalent N-ToM benchmarks have several shortcomings, including the presence of ambiguous and artificial narratives, absence of personality traits and preferences, a lack of questions addressing characters' psychological mental states, and limited diversity in the questions posed. In response to these issues, we construct \ours, a new benchmark for assessing N-ToM with (1) longer and clearer narrative stories, (2) characters with explicit personality traits, (3) actions that are triggered by character intentions, and (4) questions designed to challenge LLMs' capabilities of modeling characters' mental states of both the physical and psychological world. Using \ours, we reveal that state-of-the-art LLMs thrive at modeling certain aspects of mental states in the physical world but fall short when tracking characters' mental states in the psychological world.\footnote{Our code and data are publicly available at: 
\url{https://seacowx.github.io/projects/opentom/OpenToM.html}}

\end{abstract}

\section{Introduction}

\begin{figure} [ht!]
    \centering
    \vspace{0.75em}
    \includegraphics[width=\columnwidth]{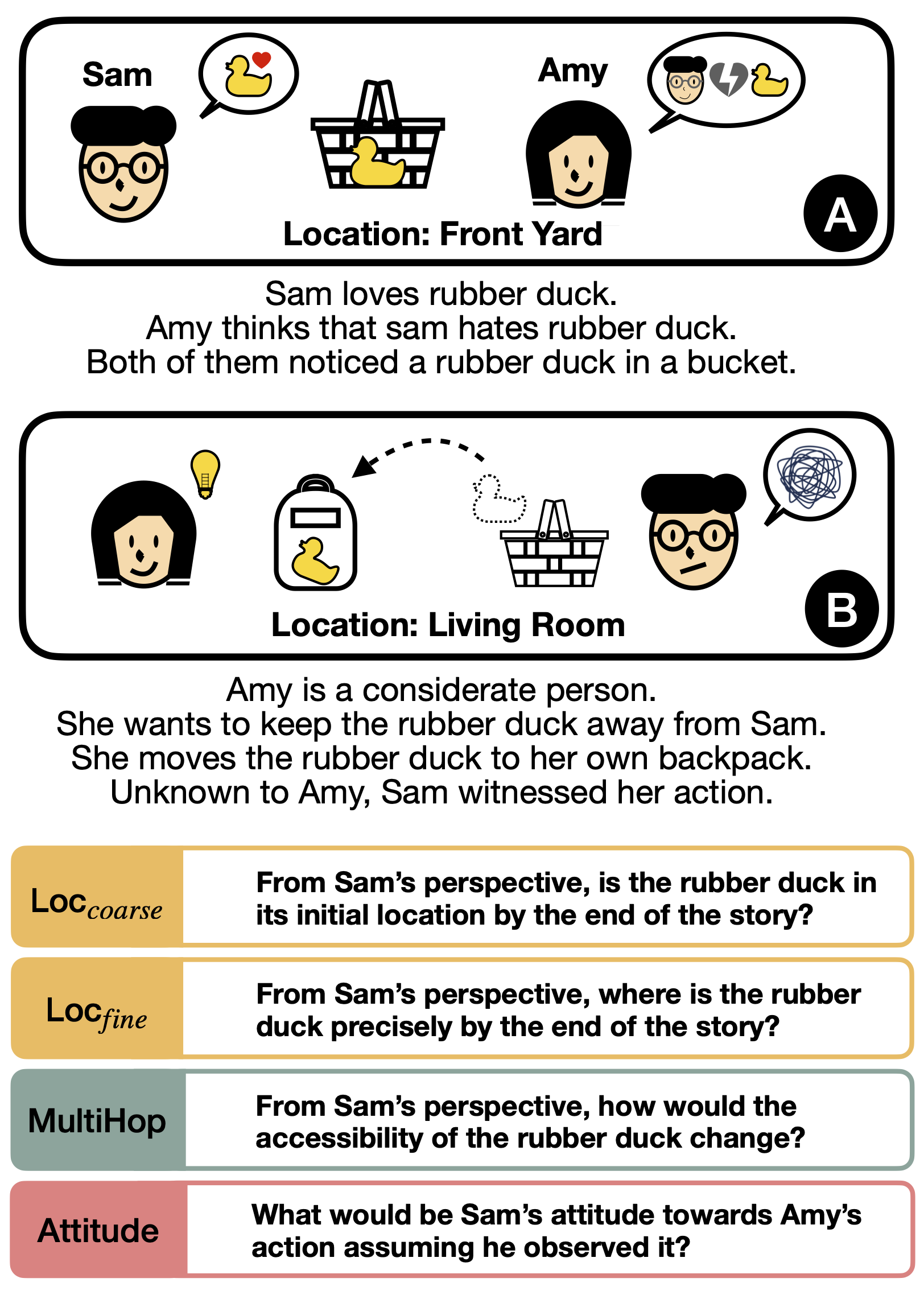}
    \caption{Illustration of a simplified story from \ours~and the corresponding first-order ToM questions. This story features two protagonists: \textit{Sam} (\textit{observer}) and \textit{Amy (mover)}; and an entity-of-interest: \textit{rubber duck}. There are two containers involved: a \textit{basket} and \textit{Amy's backpack}. Each narrative within \ours~is followed by three types of questions, namely questions regarding the location (\texttt{Loc}) of an entity, questions that involve multi-hop reasoning (\texttt{MHop}), and questions about the  characters' attitude (\texttt{Att}).}
    \label{fig:figure1}
    \vspace{-0.5em}
\end{figure}

Theory-of-Mind (ToM), the awareness that others perceive the world differently and the capability of keeping track of such differences, is at the core of social interactions \cite{premack1978does}. Studies in cognitive science have designed numerous false-belief tests to investigate human ToM capabilities \cite{premack1978does, wimmer1983beliefs, onishi200515}. One such test is the \textit{Sally-Anne Test} \cite{baron1985does}, in which Anne stealthily moves an object that is initially known to both Sally and Anne. This covert action causes Sally to have a false belief that the object is still in its initial location. Consequently, individuals taking the test are required to reason about \textit{"Where will Sally look for the object?"}

To study Neural Theory-of-Mind (N-ToM)\footnote{In this paper, we distinguish Theory-of-Mind studies between human (ToM) and artificial neural networks (N-ToM).}, machines' capabilities of performing ToM reasoning, researchers have applied human ToM tests such as the \textit{Sally-Anne Test} to benchmark Large Language Models (LLMs) \cite{le-etal-2019-revisiting,bubeck2023sparks,kosinski2023theory,Shapira2023CleverHO,Ullman2023LargeLM,wu2023hi,zhou2023far}. However, using human ToM tests for evaluating LLMs is problematic because stories in human ToM tests lack certain elements found in real-life scenarios. Specifically, the characters do not have \textbf{personality traits} or \textbf{preferences}. Additionally, their actions are \textbf{not motivated} (e.g. why would Anne want to move the object?). Furthermore, the narratives of many existing N-ToM benchmarks are generated using a template-based approach \cite{le-etal-2019-revisiting, wu2023hi, zhou2023far}, which results in overly-structured and ambiguous narratives (see Appendix~\ref{app:narrative_prompt}). The structured context makes existing benchmarks susceptible to overfitting, while the ambiguities may lead to an underestimation of a model's true N-ToM capabilities.

To this end, we introduce \textbf{Open}book-QA dataset for \textbf{ToM} (\ours). Following previous works' success in generating high-quality data using LLMs \cite{efrat2020turking, perez2022red, perez2022discovering, hartvigsen2022toxigen, west2023generative}, we generate \ours~stories using a four-stage human-in-the-loop generation pipeline (\S\ref{sec:narrative_construction}). Our pipeline includes (1) endowing characters with \textbf{preferences} and \textbf{personality traits}, (2) generating \textbf{intentions} and \textbf{the corresponding enctions} \cite{riva2011intention}, (3) constructing story plot and producing narratives using LLMs, and (4) revise and refine stories by human annotators. Based on the \ours~narratives, we formulate questions that cover characters' mental states of both \textbf{the physical world} (e.g., the location of an object) and \textbf{their psychological states} (e.g. character's attitude towards a particular action). See Figure~\ref{fig:figure1} for examples. 

We evaluate \ours~dataset on a range of LLMs including Llama2-Chat \cite{touvron2023llama}, Mixtral-8x7B-Instruct \cite{jiang2024mixtral}, GPT-3.5-Turbo \cite{openai2022chatgpt}, and GPT-4-Turbo \cite{openai2023gpt4} under a zero-shot setting. We also test two prompting techniques, namely Chain-of-Thought (CoT) \cite{wei2022chain} and SimulatedToM (SimToM) \cite{wilf2023think}. Additionally, we fine-tuned a Llama2-Chat-13B model to serve as the fine-tuning baseline. Our results show that, while fine-tuning and advanced prompting techniques improve models' N-ToM reasoning capabilities, their performance in deducing the psychological states of characters is still far from human performance (Section \ref{sec:result}). We summarize our contributions as follows:
\begin{itemize} [leftmargin=*, noitemsep]
    \item[1.] We construct \ours, a N-ToM benchmark with natural narratives, personified characters, motivated actions, and diversified questions that challenge LLMs' understanding of characters' perception of both the physical world and the psychological states. \\
    \item[2.] Using \ours, we conduct a comprehensive evaluation on representative LLMs. Our result shows a mismatch of LLMs' capability in deducing characters' mental states of the physical versus the psychological world. \\
    \item[3.] Our in-depth analysis reveals LLMs' shortcomings in N-ToM including unfaithfulness in N-ToM reasoning, sensitivity to narrative length and character roles, and lack of understanding of characters' psychological perception.
\end{itemize}

\begin{figure*}
    \centering 
    \includegraphics[width=\textwidth]{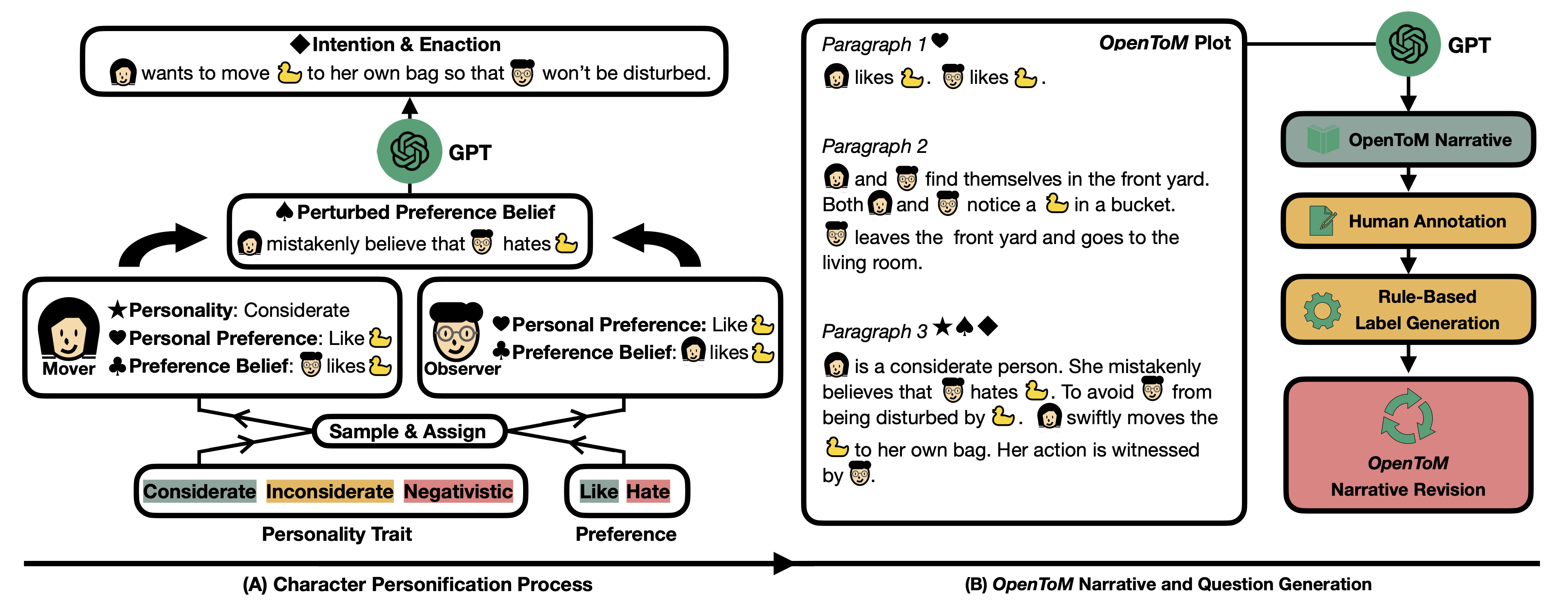}
    \caption{The data generating process of \ours~dataset. Using the story in Figure~\ref{fig:figure1} as an example, the features created in the personification process are shown in Part (A), which include character preference ($\varheart$), belief of the other character's preference ($\clubsuit$), the perturbed \textit{mover}'s preference belief ($\spadesuit$), the \textit{mover}'s personality trait ($\bigstar$), and the \textit{mover}'s intention and action ($\blacklozenge$). The usage of these information in the \ours~plot are shown in Part (B) next to the paragraph indicator. See Appendix~\ref{app:annotation} for detailed description of the \textit{Human Annotation} and \textit{Rule-Based Label Generation} process.}
    \label{fig:data_gen}
    \vspace{-0.5em}
\end{figure*}

\vspace{-0.5em}
\section{The \ours~Dataset}
The omissions of characters' personality, intention, and enaction in existing N-ToM benchmarks makes it difficult to construct questions that inquire \textbf{characters' mental states of the psychological world}. To address this, each of the characters in \ours~stories is \textbf{personified} and \textbf{acts with an intention} (Appendix\ref{app:personification}). Recognizing that LLMs are good at utilizing spurious correlations such as lexical overlaps \cite{Shapira2023CleverHO}, we take extra effort in mitigating the potential spurious cues in \ours~stories (\S\ref{sec:spurious_cue}).

\subsection{\ours~Construction}
\label{sec:narrative_construction}
A typical \ours~story consists of two protagonists, an entity-of-interest (referred to as the "entity" henceforth), and several locations and containers. Of the two protagonists, one is assumed as the role of the \textit{mover}, who carries out actions on the entity, and another is the \textit{observer}, who may or may not witness these actions (see Figure~\ref{fig:figure1}).

As shown in Figure~\ref{fig:data_gen}, the data generating process consists of two main stages, namely the \textit{Character Personification Process} followed by the \textit{Narrative and Question Generation Process}. We start the anthropomorphism process by assigning a personality trait and personal preference to each character. Specifically, the personality traits are sampled from three candidates (see Appendix~\ref{app:personification}, and Algorithm~\ref{alg:generate_preference}) and the preference is randomly chosen from binary options. To mitigate spurious correlation, we create false beliefs on characters' perception of each other's personal preferences by randomly flipping the preference label (see Section~\ref{sec:spurious_cue}). Using the sampled personal preferences, personality traits, and a world state initialized from ToMi \cite{le-etal-2019-revisiting}, we prompt GPT-3.5-Turbo to generate the \textit{mover}'s intention and enactions. The enaction results in world state changes, which are used to construct the final world state. We use this information to draft a story plot, refer to as the \ours~plot.

A \ours~plot consists of three paragraphs. The first paragraph illustrate the characters' personal preferences and their beliefs about each other's preferences. The second paragraph serves as the prologue, which depicts the initial world state and some preceding events involving the two characters. The last paragraph describes the main event, which includes the \textit{mover}'s personality, the \textit{mover}'s intention, and their subsequent action. It is worth noting that, in order to reduce ambiguity, we  explicitly include information regarding whether the \textit{observer} perceived the \textit{mover}'s action. We carefully designed the plot as well as the narrative generating process so that the \textit{observer}'s mental activity is excluded from the final \ours~narrative while ensuring that the \textit{observer}'s perception of the main event is mentioned.

After generating the \ours~narratives, we classify the corresponding ToM questions into two categories, those requiring human annotation and those that can be automatically annotated using human-defined labels combined with first-order logic (see Appendix~\ref{app:annotation}). In the final stage of data generation, we conduct a round of quality inspection. Specifically, we examine each narrative to ensure that (1) the answers to the ToM questions are not directly given in the narrative, (2) The narrative content aligns with commonsense knowledge, and (3) there is no significant lexical overlaps between the narrative and the corresponding ToM questions (as discussed in Section~\ref{sec:spurious_cue}).


\subsection{\ours~Overview}

Overall, \ours~contains 696 narratives. We first produce 596 narratives with GPT-3.5-Turbo\footnote{We used the GPT-35-1106 checkpoint through Microsoft Azure OpenAI service. All \ours~narratives are generated in December 2023. We also tested with GPT-4-1106 and obtained narratives of similar quality. Hence we choose GPT-3.5-Turbo for its lower cost.} using the pipeline shown in Figure~\ref{fig:data_gen}. In addition, we sample 100 existing \ours~plots and produce extra-long narratives (\ours-L) using GPT-4-Turbo\footnote{We used the GPT-4-1106 checkpoint through Microsoft Azure OpenAI service. All \ours-L narratives are generated in December 2023.}. To elicit the unique N-ToM challenges posted by our \ours~benchmark, we compare \ours~with established N-ToM benchmarks in Table~\ref{tab:benchmark_comparison}. See Appendix~\ref{app:data_statistics} for detailed statistics of the \ours~benchmark.

\subsection{Task Formulation}
\label{sec:task_formulation}

We formulate all \ours~questions as binary or ternary classification tasks (see Figure~\ref{fig:label_dist} for detailed label space and label distributions). Formally, given a complete narrative $\mathcal{N}_{comp}$, a set of answers $\mathcal{A}$, a character $c$, and a character-centric question $q_{c}$. A model is to first deduce the information accessible to character $c$, denoted as $\mathcal{N}_{c}$, and then answer the question. The process of extracting a character-centric narrative $\mathcal{N}_{c}$ can be made explicit, as in ~\citet{wilf2023think}, or latent, as is common in most ToM evaluations. In general, the \ours~task can be formulated as follows:
\[
a^{*}_{c} = \text{argmax}_{a \in \mathcal{A}} \mathbb{P}\big(a \mid \mathds{1}_{expl}\cdot\mathcal{N}_{c}, \mathcal{N}_{comp}, q_{c}\big)
\]
where $\mathds{1}_{expl}$ is an indicator function that returns 1 if the character-centric narrative is explicitly provided and 0 otherwise.

\begin{table} [t]
    \centering
    \resizebox{\columnwidth}{!}{
    \begin{tabular}{c|ccccc|cc}
        \toprule
        \multicolumn{4}{l}{{\color{yellow} \faUserFriends} : Social Commonsense} &
        \multicolumn{4}{l}{{\color{blue} \faAtom} : Physical ToM} \\
        \multicolumn{4}{l}{{\color{brain} \faBrain} : Psychological ToM} &
        \multicolumn{4}{l}{{\color{green} \faUser} : Personified Character} \\
        \multicolumn{4}{l}{{\color{narrative} \faFile*} : Number of Narratives} &
        \multicolumn{4}{l}{{\color{red} \faPenNib} : Average Token Count} \\
        \multicolumn{4}{l}{\faLink: Structured Narrative} &
        \multicolumn{4}{l}{\faUnlink: Unstructured Narrative} \\
        \toprule
        \toprule
        & Narrative & 
        {\color{yellow} \faUserFriends} & 
        {\color{blue} \faAtom} & 
        {\color{brain} \faBrain} & 
        {\color{green} \faUser} & 
        {\color{narrative} \faFile*} & 
        {\color{red} \faPenNib} 
        \\
        \midrule
        ToMi & \faLink & \ding{56} & \ding{52} & \ding{56} & \ding{56} & 999 & 44.6 \\
        $\text{T4D}^{a}$ & \faLink & \ding{56} & \ding{52} & \ding{56} & \ding{56} & $\sim$500 & $\sim$50 \\
        Adv-CSFB & \faLink & \ding{56} & \ding{52} & \ding{56} & \ding{56} & 40 & 70.8 \\
        Hi-ToMi & \faLink & \ding{56} & \ding{52} & \ding{56} & \ding{56} & 1200 & 213.68 \\
        Big-ToMi & \faUnlink & \ding{56} & \ding{52} & \ding{56} & \ding{52} & 3000 & 69.9 \\
        FANToM & \faUnlink & \ding{56} & \ding{52} & \ding{56} & \ding{56} & 254 & 1020.0 \\
        $\text{G-DRAGON}^{b}$ & $\text{PBP}^{c}$ & \ding{56} & \ding{56} & \ding{56} & \ding{56} & $\sim$800K & $\sim$72.5 \\
        FauxPas-EAI & \faUnlink & \ding{52} & \ding{52} & \ding{52} & \ding{52} & 44 & 60.5 \\
        \midrule 
        \midrule 
        \ours & \faUnlink & \ding{52} & \ding{52} & \ding{52} & \ding{52} & 596 & 194.3 \\
        \ours-L & \faUnlink & \ding{52} & \ding{52} & \ding{52} & \ding{52} & 100 & 491.6 \\
        \bottomrule
        \multicolumn{8}{l}{(a, b) Not open-sourced. The number of narratives and average tokens are} \\[-0.3em]
        \multicolumn{8}{l}{estimated according to \citet{zhou2023far} and \citet{zhou-etal-2023-cast}.} \\
        \multicolumn{8}{l}{(c) PBP: Play-By-Post game play data of Dungeons\&Dragons. See} \\[-0.3em]
        \multicolumn{8}{l}{\citet{zhou-etal-2023-cast} for details.} \\
    \end{tabular}
    }
    \vspace{-0.75em}
    \caption{Comparison of \ours~benchmark with existing N-ToM datasets. In the header, \textit{Physical ToM} and \textit{Psychological ToM} refers to testing ToM capabilities in characters' mental states of the physical world and the psychological world respectively.}
    \vspace{-1em}
    \label{tab:benchmark_comparison}
\end{table}

\subsection{Question Genres}
\label{sec:question_genres}

Each of \ours~stories is accompanied by 23 questions that cover both \textit{first-order} ToM and \textit{second-order} ToM. \textit{First-order} ToM questions, which directly ask about a character's perception of the world, is illustrated in the bottom of Figure~\ref{fig:figure1}. \textit{Second-order} ToM questions inquire about a character's belief of another character's mental state. For instance, a second-order ToM question based on the story in Figure~\ref{fig:figure1} could be "\textit{From Sam's perspective, does Amy think the rubber duck is in its initial location?}". Overall, \ours~questions can be summarized into the following 3 genres:\vspace{0.75em} \\
\textbf{Location (\texttt{Loc})} questions are concerned with the characters' perception of the entity's location. In \ours, we create two versions of location questions,  $\texttt{Loc}_{coarse}$ and $\texttt{Loc}_{fine}$. $\texttt{Loc}_{coarse}$ asks about the character's perception of whether an entity is at its initial location, while $\texttt{Loc}_{fine}$ inquires about the entity's explicit location (see Figure~\ref{fig:figure1} for an example). By doing so, we wish to mitigate the impact of location granularity (Appendix~\ref{app:location_granularity}) and assess the model's faithfulness in answering this type of questions (\S\ref{sec:location_faithfulness} and Appendix~\ref{app:location_granularity}). \vspace{0.75em}\\
\textbf{Multi-Hop (\texttt{MHop})} questions are composed by adding an additional reasoning hop on top of the \texttt{Loc} questions. Specifically, we inquire about changes in the \textit{fullness} of the containers and the \textit{accessibility} of the entity (see Figure~\ref{fig:figure1} for an example), all of which demand 3-hop reasoning (illustrated in Appendix~\ref{app:multihop_demo}).  

To address the lack of \textbf{social commonsense} in previous N-ToM benchmarks \cite{ma-etal-2023-towards-holistic}, we have devised the \textit{accessibility} questions specifically for testing LLMs' understanding of social norms. Taking the \texttt{MHop} question in Figure~\ref{fig:figure1} as an example, in attempting to answer this question, a model needs to first reason whether the character knows about the rubber duck's movement. The need for social commonsense comes in the next reasoning hop. Assuming the model is aware that the rubber duck is in Amy's backpack, it must grasp the social commonsense that others shall not take things from Amy's backpack without permission. Therefore, a model with adequate social intelligence shall respond with "\textit{less accessible}" 
\vspace{0.75em}\\
\textbf{Attitude (\texttt{Att})} questions are designed to challenge LLMs' capability to interpret a character's psychological mental state. Specifically, LLMs are required to deduce the \textit{observer}'s potential attitude towards the \textit{mover}'s action (see Figure~\ref{fig:figure1} for an example). As discussed in \S\ref{sec:spurious_cue}, the crux of solving \textit{attitude} questions is to first identify the information accessible to the \textit{observer} and then use social commonsense to infer the \textit{attitude}. In \ours, of all the knowledge related to the \textit{observer}'s \textit{attitude}, only the \textit{observer}'s own preference towards the entity and the \textit{mover}'s action are accessible to the \textit{observer} (see Figure~\ref{fig:spurious_cue}). Therefore, \ours~stories are carefully crafted so that LLMs may not succeed by leveraging information inaccessible to the \textit{observer} (\S\ref{sec:spurious_cue}). 

Human's \textit{attitude} is subjective and multifaceted \cite{zhan-etal-2023-evaluating}, we reduce such complexity by maximizing the contrast between the \textit{observer}'s preference and the \textit{mover}'s action. In the story of Figure~\ref{fig:figure1}, Amy moves Sam's favorite rubber duck into her own backpack. The substantial disparity between Sam's fondness of the rubber duck and Amy's seemingly selfish act will likely cause Sam to have a negative attitude towards Amy's action. Our data validation study (\S\ref{sec:data_validation}) shows the effectiveness of this approach.

\subsection{Mitigating Spurious Correlation}
\label{sec:spurious_cue}
We take measures to mitigate spurious correlation in all questions. Fixing the \texttt{Loc} and \texttt{MHop} questions can be done by revising narratives based on keywords. We identify \ours~narratives that contain phrases which have substantial lexical overlap with the questions or those that provide shortcuts for answering them (Appendix~\ref{app:spurious_words}). We manually revise such narratives to reduce the reporting bias, resulting in revisions for 17.8\% of the \ours~narrative drafts.

\begin{figure} [t]
    \centering
    \includegraphics[width=\columnwidth]{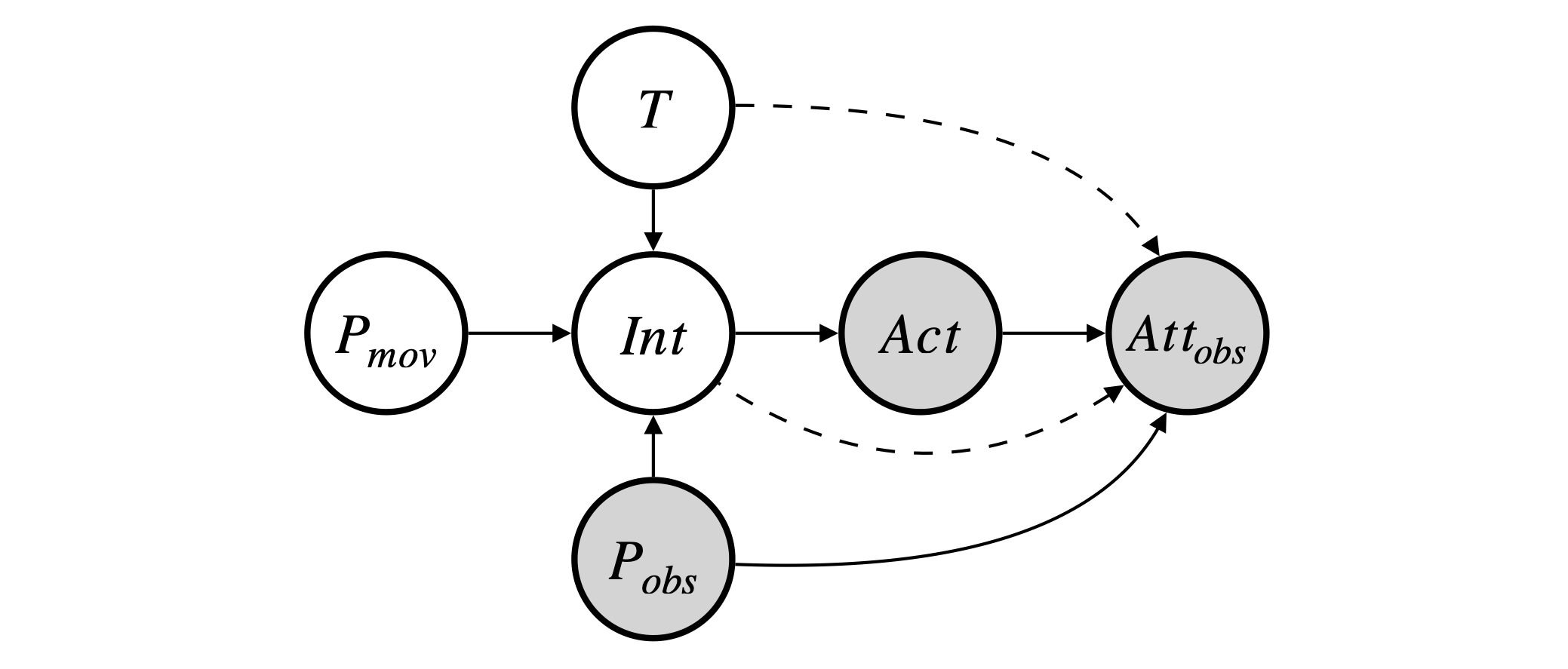}
    \caption{A Bayesian Network representation of the dependencies among preference ($P$), personality trait ($T$), intention ($Int$), action ($Act$), and attitude ($Att$). The causal relations are represented by solid arrows. The spurious correlations are represented by dashed arrows. The grey-shaded variables are observable by the \textit{observer} and the unshaded variables are latent to the \textit{observer}.}
    \vspace{-1em}
    \label{fig:spurious_cue}
\end{figure}

To elicit the potential spurious cues in \textit{Attitude} questions, we define the enaction process as a Bayesian network \cite{riva2011intention, baker2011bayesian} (Figure~\ref{fig:spurious_cue}). Firstly, the intention of the \textit{mover} ($Int$) originates from their preference ($P_{mov}$), their personality trait ($T$), and, optionally, the \textit{observer}'s preference ($P_{obs}$). This process is latent for the \textit{observer}-- the only observable variables are their own preference ($P_{obs}$) and the action ($Act$). Employing the $do$-calculus notation from \citet{pearl1995causal}, solving the \textit{attitude} question is equivalent to solving the following problem
\[
att^{*} = \text{argmax}_{att \in Att_{obs}} \mathbb{P}(att \mid do(act), P_{obs})
\]
where $att$ is an instantiation of the \textit{observer}'s potential attitudes, $Att_{obs}$. Overall, we identify two types of potential spurious cues, (1) model $\mathbb{P}(att \mid Int)$ or (2) model $\mathbb{P}(att \mid T)$, as shown in Figure~\ref{fig:spurious_cue}. We show that addressing these two spurious correlations concurrently can be achieved by adjusting the \textit{mover}'s beliefs regarding the \textit{observer}'s preference (see Appendix~\ref{app:spurious_cue} for details).

\begin{table*}[t]
    \centering
    \resizebox{\linewidth}{!}{
    \begin{tabular}{ c | c | c c | c c c c c c | c }
    \toprule
        & \textbf{Human} & \multicolumn{2}{c|}{\textbf{Naive Baseline}} & \multicolumn{6}{c|}{\textbf{Large Language Models}} & \textbf{FT.} \\[0.5em]
        &  & Ran. & Maj. & \multicolumn{3}{c}{Llama2-Chat} & Mixtral-Instruct & GPT-3.5-Turbo & GPT-4-Turbo & Llama2 \\
    \# Params & --- & --- & --- & 7B & 13B & 70B & 8x7B & --- & --- & 13B \\
    \midrule
        $\texttt{Loc}_{c}$ (F) & 0.990 & 0.491 & 0.416 & $0.290_{\pm0.045}$ & $0.391_{\pm0.022}$ &$0.413_{\pm0.016}$ & $0.512_{\pm0.044}$ & $0.439_{\pm0.025}$ & \boldmath{$0.643_{\pm0.061}$} & 0.978  \\
    \midrule
        $\texttt{Loc}_{c}$ (S) & 0.993 & 0.467 & 0.381 & \boldmath{$0.462_{\pm0.069}$} & $0.355_{\pm0.043}$ & $0.280_{\pm0.028}$ & $0.294_{\pm0.025}$ & $0.323_{\pm0.039}$ & $0.442_{\pm0.044}$ & 0.749 \\
    \midrule
        $\texttt{Loc}_{f}$ (F) & 0.990 & 0.000 & 0.003 & $0.404_{\pm0.029}$ & \boldmath{$0.545_{\pm0.023}$} & $0.534_{\pm0.023}$ & $0.399_{\pm0.015}$ & $0.515_{\pm0.012}$ & $0.507_{\pm0.010}$ & 0.600 \\
    \midrule
        $\texttt{Loc}_{f}$ (S) & 0.993 & 0.000 & 0.002 & $0.245_{\pm0.015}$ & \boldmath{$0.301_{\pm0.006}$} & $0.223_{\pm0.023}$ & $0.211_{\pm0.011}$ & $0.286_{\pm0.006}$ & $0.269_{\pm0.004}$ & 0.495 \\
    \midrule
        $\texttt{MHop}$ (F) & 0.855 & 0.345 & 0.182 & $0.322_{\pm0.026}$ & $0.301_{\pm0.023}$ & $0.501_{\pm0.026}$ & $0.556_{\pm0.026}$ & $0.468_{\pm0.029}$ & \boldmath{$0.658_{\pm0.034}$} & 0.936 \\
    \midrule
        $\texttt{MHop}$ (S) & 0.770 & 0.323 & 0.219 & $0.211_{\pm0.024}$ & $0.229_{\pm0.037}$ & $0.434_{\pm0.048}$ & $0.474_{\pm0.025}$ & $0.334_{\pm0.025}$ & \boldmath{$0.637_{\pm0.034}$} & 0.784 \\
    \midrule
        $\texttt{Att}$ & 0.862 & 0.328 & 0.174 & $0.240_{\pm0.027}$ & $0.375_{\pm0.031}$ & $0.415_{\pm0.051}$ & $0.476_{\pm0.041}$ & $0.410_{\pm0.021}$ & \boldmath{$0.544_{\pm0.060}$} & 0.547 \\
    \bottomrule
    \end{tabular}}
    \vspace{-5pt}
    \caption{Evaluation results in Macro-averaged F1 scores of the \ours~dataset. Location subscripts, $c$ and $f$, represents \textit{coarse} and \textit{fine} respectively. The capital \textit{F} and \textit{S} in the parenthesis represent \textit{first-order ToM} and \textit{second-order ToM}. The naive baselines include a random guess (Ran.) and a majority (Maj.) baseline. The finetuning baseline (FT.) is a Llama2-Chat 13B model finetuned following the configuration in Appendix~\ref{app:ft_config}.}
    \vspace{-0.5em}
    \label{tab:main_results}
\end{table*}

\subsection{Dataset Validation}
\label{sec:data_validation}
To verify the human performance and agreement on the \ours~dataset, we sampled 100 narratives, each of which contains 5 sampled questions covering all 3 question genres asked for both \textit{first-order} and \textit{second-order} ToM (see Figure~\ref{fig:annotation} for a demonstration of the data annotation interface). This set of \ours~data are annotated independently by 3 annotators. The inter-annotator agreement is reflected through the macro-averaged F1 score (Table~\ref{tab:main_results}), which is computed as the arithmetic mean of the pairwise agreement scores (see Appendix~\ref{app:agreement} for detailed statistics). The agreement scores demonstrate that the \ours~questions contain minimal subjectivity and align well with the collective judgement of human annotators.

\section{Experiments}
Following the convention of previous N-ToM studies, we focus on evaluating zero-shot performance of LLMs \cite{Shapira2023CleverHO, kim-etal-2023-fantom, sclar-etal-2023-minding, zhou2023far}.
\subsection{Baseline Models}
We evaluate the \ours~tasks using 6 representative LLMs, namely the Llama2-Chat models (7B, 13B, and 70B) \cite{touvron2023llama}, the Mixtral-8x7B-Instruct model \cite{jiang2024mixtral}, and the GPT-3.5-Turbo and GPT-4-Turbo\footnote{We use the 1106 checkpoints of the GPT-3.5-Turbo and GPT-4-Turbo models. The experiments are run between December 2023 and January 2024 using API provided by Microsoft Azure OpenAI Studio \url{https://oai.azure.com/}.} models \cite{openai2022chatgpt, openai2023gpt4}. We also fine-tuned a Llama2-Chat 13B model (Appendix~\ref{app:ft_config}). See Appendix~\ref{app:eval_models} for detailed description of the models.

\subsection{Prompting Techniques}
In addition to the vanilla prompting, we experiment with two additional prompting techniques, namely Chain-of-Thought (CoT) \cite{wei2022chain} and SimulatedToM (SimTom) \cite{wilf2023think}. CoT prompting is widely used in reasoning tasks. It demands LLMs to explicitly generate its step-by-step reasoning process. SimToM prompting is specifically designed to aid N-ToM tasks, which asks LLMs to first generate a character-centric narrative, $\mathcal{N}_{c}$, and then answer character-specific questions.

\subsection{Overall Results}
\label{sec:result}
As all the \ours~questions are formulated as binary or ternary classification tasks and considering that the labels are not uniformly distributed (Figure~\ref{fig:label_dist}), we evaluate model performance using the macro-averaged F1 scores (referred to as F1 scores henceforth).

To evaluate the consistency of LLMs' performance, we randomly sample 50 narratives for each round of evaluation and repeat this process for 5 times for each model. We compute the mean and the standard deviation of the F1 scores, which are reported in Table~\ref{tab:main_results} (See Table~\ref{tab:detailed_main_results} for more detailed results. See Table~\ref{tab:mhop_breakdown} for the breakdown of LLMs' performances on \texttt{MHop} questions). Overall, we see that GPT-4-Turbo outperforms other models on $\texttt{Loc}_{coarse}$ (first-order), \texttt{MHop}, and \texttt{Att} questions by a large margin. However, we are surprised to see that Llama2-Chat-7B performs the best in answering second-order $\texttt{Loc}_{coarse}$. However, due to the high unfaithful rate shown in later studies (\S\ref{sec:location_faithfulness} and Table~\ref{tab:faithfulness_numbers}), achieving the highest score does not necessarily imply that Llama2-Chat-7B is more capable in N-ToM. In addition, it is interesting to see that, while GPT-4-Turbo leads in most question genres by a large margin, its capability of answering the $\texttt{Loc}_{fine}$ questions is not on par with Llama2-Chat-13B, 70B, or GPT-3.5-Turbo.

Through the fine-tuning model, it becomes evident that the $\texttt{Loc}_{coarse}$ and $\texttt{MHop}$ questions are easier to learn, as their F1 scores improved dramatically. On the other hand, the $\texttt{Loc}_{fine}$ and \texttt{Att} questions pose greater challenges as the F1 score of the fine-tuned model only have limited improvement. 

CoT prompting brings significant performance gains to all models on $\texttt{Loc}_{coarse}$ and $\texttt{MHop}$ questions. However, the improvements in answering \texttt{Att} questions are marginal and the performance on $\texttt{Loc}_{fine}$ questions declines. In the case of SimToM prompting, the results for the Mixtral model are mixed. SimToM improves the f1 score of \texttt{MHop} questions, but its performance on other question types is either degraded or negligible. For GPT models, SimToM consistently brings performance gains in $\texttt{Loc}_{coarse}$ questions. However, for other question genres, the effect of SimToM is mixed.

In terms of the length of the narrative, results on \ours-L show that ToM in longer narratives are generally harder to trace. Please see Appendix~\ref{app:narrative_length} for detailed results and analysis. 

\begin{table} [h]
    \centering
    \resizebox{\linewidth}{!}{
    \begin{tabular} {P{0.01\columnwidth} c | c c | c c | c c | P{0.25cm}}
        \toprule
        & Question & \multicolumn{2}{c|}{Mixtral} & \multicolumn{2}{c|}{GPT-3.5-Turbo} & \multicolumn{2}{c|}{GPT-4-Turbo} & HL \\
        & & F1 & $\Delta$F1 & F1 & $\Delta$F1 & F1 & $\Delta$F1 & \\
        \midrule 
        \multirow{7}{*}{\rotatebox[origin=c]{90}{CoT}} 
        & $\texttt{Loc}_{c} (F)$ & 0.784* & \tableinc{+0.272} & 0.587* & \tableinc{+0.148} & \boldmath{$0.942$}* & \tableinc{+0.299} & \ding{52} \\ 
        & $\texttt{Loc}_{c} (S)$ & 0.539* & \tableinc{+0.245} & 0.457* & \tableinc{+0.134} & \boldmath{$0.828$}* & \tableinc{+0.386} & \ding{56} \\
        & $\texttt{Loc}_{f} (F)$ & 0.301* & \tabledec{- 0.098} & \boldmath{$0.469$}* & \tabledec{- 0.046} & 0.450* & \tabledec{- 0.057} & \ding{56} \\
        & $\texttt{Loc}_{f} (S)$ & 0.180* & \tabledec{- 0.031} & \boldmath{$0.240$}* & \tabledec{- 0.046} & 0.187* & \tabledec{- 0.082} & \ding{56} \\
        & $\texttt{MHop} (F)$ & 0.610* & \tableinc{+0.054} & 0.547* & \tableinc{+0.079} & \boldmath{$0.835$}* & \tableinc{+0.177} & \ding{52} \\
        & $\texttt{MHop} (S)$ & 0.551* & \tableinc{+0.077} & 0.414* & \tableinc{+0.080} & \boldmath{$0.755$}* & \tableinc{+0.118} & \ding{52} \\
        & $\texttt{Att}$ & 0.519* & \tableinc{+0.043} & 0.446* & \tableinc{+0.036} & \boldmath{$0.580$}* & \tableinc{+0.036} & \ding{56} \\
        \midrule
        \multirow{7}{*}{\rotatebox[origin=c]{90}{SimToM}} 
        & $\texttt{Loc}_{c} (F)$ & 0.414* & \tabledec{- 0.098} & 0.635* & \tableinc{+0.196} & \boldmath{$0.838$}* & \tableinc{+0.195} & \ding{56} \\
        & $\texttt{Loc}_{c} (S)$ & 0.290 & \tablesame{- 0.004} & 0.400* & \tableinc{+0.077} & \boldmath{$0.685$}* & \tableinc{+0.243} & \ding{56} \\ 
        & $\texttt{Loc}_{f} (F)$ & 0.352* & \tabledec{- 0.047} & \boldmath{$0.518$}* & \tablesame{+0.003} & 0.485* & \tabledec{- 0.022} & \ding{56} \\
        & $\texttt{Loc}_{f} (S)$ & 0.206* & \tablesame{- 0.005} & \boldmath{$0.261$}* & \tabledec{- 0.025} & 0.217* & \tabledec{- 0.079} & \ding{56} \\
        & $\texttt{MHop} (F)$ & 0.650* & \tableinc{+0.094} & 0.536* & \tableinc{+0.068} & \boldmath{$0.720$}* & \tableinc{+0.062} & \ding{56} \\
        & $\texttt{MHop} (S)$ & 0.514* & \tableinc{+0.040} & 0.350* & \tableinc{+0.016} & \boldmath{$0.631$}* & \tablesame{- 0.006} & \ding{56} \\
        & $\texttt{Att}$ & 0.404* & \tabledec{- 0.072} & 0.416 & \tablesame{+0.006} & \boldmath{$0.488$}* & \tabledec{- 0.056} & \ding{56} \\
        \bottomrule
    \end{tabular}
    }
    \vspace{-5pt}
    \caption{Macro F1 score of \ours~dataset evaluated using CoT and SimToM prompting with relative \tableinc{performance gain}, \tabledec{performance degradation}, or \tablesame{equal performance} ($\Delta\text{F1} < 0.010$). "*" indicates statistical significance under the Two-sample T test with a level of significance of $\alpha=0.05$. The score of the best performing model on each task is bolded. HL (human level) indicates whether the performance of the best model is on par with human performance (within a margin of 0.050).}
    \vspace{-0.5em}
    \label{tab:fancy_prompting}
\end{table}

\section{Detailed Result Analysis}
To further investigate LLMs' N-ToM capabilities, we conduct in-depth analysis on LLMs' faithfulness in answering $\texttt{Loc}_{coarse}$ and $\texttt{Loc}_{fine}$ questions (\S\ref{sec:location_faithfulness}), performance discrepancy of modeling the mental states of different character roles (\S\ref{sec:mover_vs_observer}), and lack of capability in modeling characters' mental state of the psychological world (\S\ref{sec:attitude}).

\subsection{Faithfulness in \texttt{Loc} Questions}
\label{sec:location_faithfulness}

As mentioned in \S\ref{sec:question_genres}, we create two types of \texttt{Loc} questions differ in granularity. In principle, $\texttt{Loc}_{coarse}$ serves as a prerequisite for answering $\texttt{Loc}_{fine}$ questions. For instance, if a person believes that the entity is not in its initial location (i.e. $\texttt{Loc}_{coarse}$), then they should maintain this belief when deducing its precise location (i.e. $\texttt{Loc}_{fine}$). We conduct two experiments to examine LLMs' faithfulness\footnote{We follow the definition of "faithfulness" from \citet{jacovi-goldberg-2020-towards}, which is \textit{"the true reasoning process behind the model’s prediction"}. We regard the model as unfaithful when its \textit{true reasoning process} deviate from that of human.} in answering the $\texttt{Loc}$ questions. In the \textit{Joint} approach, we present LLMs with $\texttt{Loc}_{coarse}$ which is immediately followed by $\texttt{Loc}_{fine}$ in the same session. In the \textit{Separate} approach, we prompt LLMs with each \texttt{Loc} question individually. 

\begin{figure} [t]
    \centering
    \includegraphics[width=\columnwidth]{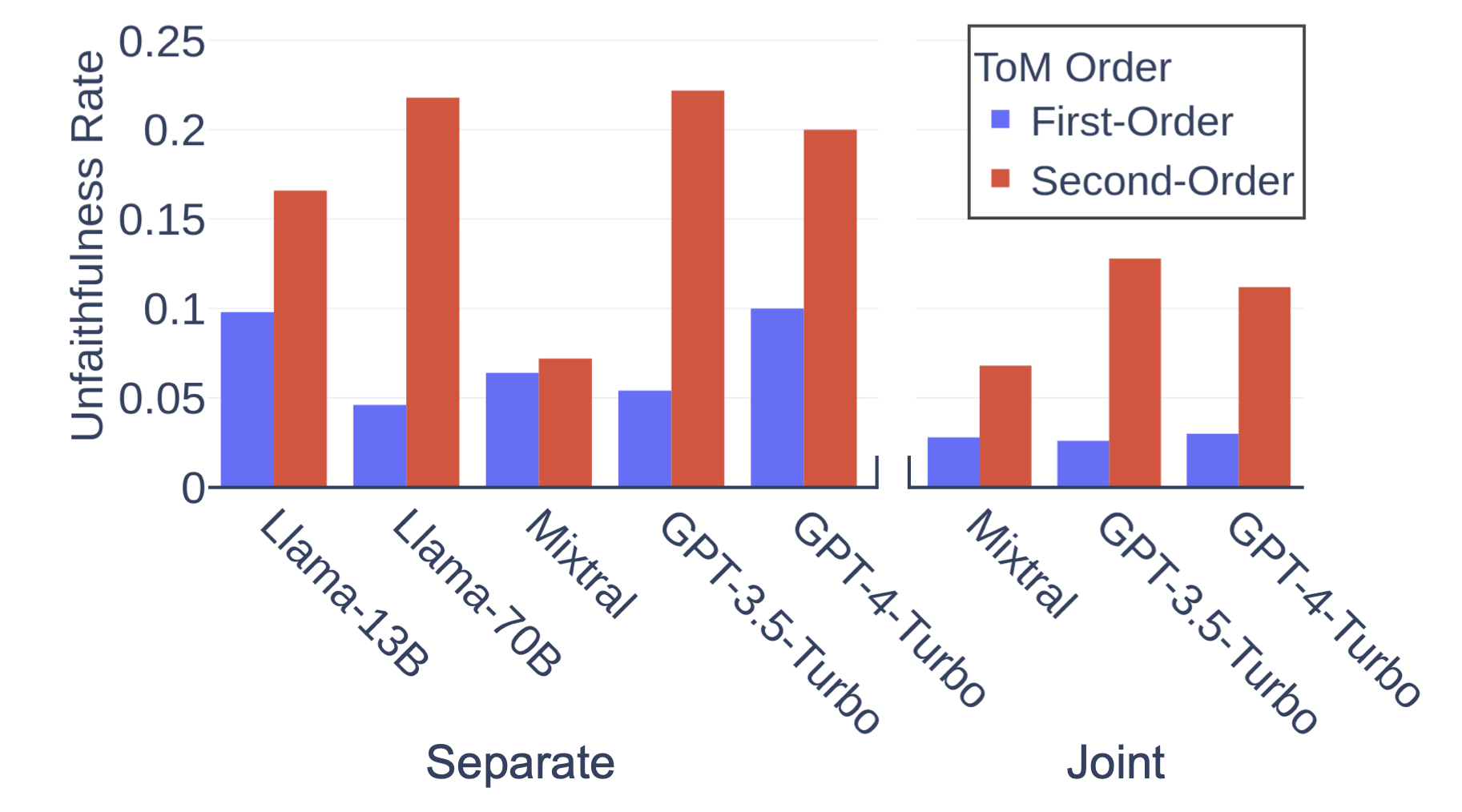}
    \caption{Faithfulness of LLMs in answering \texttt{Loc} questions. The x-axis displays the evaluation model and the y-axis displays the \textit{Unfaithful Rate}.}
    \label{fig:faithfulness}
    \vspace{-0.5em}
\end{figure}

We consider a model to be \textit{Unfaithful} if it gives contradictory answers in the $(\texttt{Loc}_{fine}, \texttt{Loc}_{coarse})$ pair of questions. To quantify this, we compute the \textit{Unfaithful Rate} for each model, which is the ratio of unfaithful pairs to the total number of pairs, as shown in Figure~\ref{fig:faithfulness}.

We see that each model's unfaithful rate is lower when answering first-order ToM questions. This is likely due to their relative simplicity comparing to the second-order questions. Further, we see that, for the GPT models, the \textit{Joint} approach yields lower \textit{Unfaithful Rate} than the \textit{Separate} approach. This improvement may attribute to having access to the previous answer in the context. For Mixtral model, however, the same trend is only observed for the first-order questions. As delving into the reason behind this trend is beyond the scope of this paper, we leave it as future work. Detailed evaluation results are shown in Appendix~\ref{app:faithfulness_numbers}.

\subsection{Performance Gap in Character Roles}
\label{sec:mover_vs_observer}
Previous works discovered that LLMs are more capable of answering questions related to the protagonist \cite{sap-etal-2022-neural, Shapira2023CleverHO}, which is likely due to them receiving more descriptions regarding their mental states \cite{grosz-etal-1995-centering}. In \ours, we consciously avoid such a reporting bias (\S\ref{sec:spurious_cue}). However, apart from the bias towards the protagonists, we observe that there exists another performance discrepancy in modeling the mind of characters of different roles. In \ours, the roles are \textit{mover} and \textit{observer}.

To demonstrate the performance gap between the \textit{mover}'s and the \textit{observer}'s perception, we compute difference in F1 scores between the models' performance on \textit{mover}-centric questions and \textit{observer}-centric questions (Table~\ref{tab:mover_vs_observer}). 

For second-order \texttt{Loc} questions, the majority of LLMs perform worse when modeling the \textit{mover}'s mental state. This is likely due to the long distance between the description of the \textit{mover}'s action and whether the \textit{observer} witnessed the action (see an examples in Appendix~\ref{app:opentom_examples}). Such distant information make it difficult for LLMs to establish a connection. Hence, deducing the \textit{mover}'s perception of the \textit{observer}'s mental state becomes more challenging.

For \texttt{MHop} questions, all LLMs perform better when modeling the \textit{mover}'s mental states. When answering first-order \texttt{Mhop} questions, models' burden for deciding whether the \textit{mover} observed their own action is alleviated. In the case of second-order \texttt{MHop} questions, the performance discrepancy is likely due to the explicit mention of the \textit{mover}'s intention. These intentions often involve the \textit{mover}'s perception of the consequences of their actions on the \textit{observer}, which greatly reduces the complexity of modeling the \textit{mover}'s perception of the \textit{observer}'s mental state. 

\begin{table} [t]
    \resizebox{\columnwidth}{!}{
        \begin{tabular}{c | c | c | c | c | c}
            \toprule
            & Llama-13B & Llama-70B & Mixtral & GPT-3.5T & GPT-4T \\
            \midrule
            $\texttt{Loc}_c$ (F) & \tableinc{+0.169} & \tableinc{+0.711} & \tableinc{+0.606} & \tableinc{+0.686} & \tableinc{+0.464} \\
            $\texttt{Loc}_c$ (S) & \tableinc{+0.047} & \tabledec{- 0.035} & \tabledec{- 0.040} & \tabledec{- 0.029} & \tableinc{+0.129} \\
            $\texttt{Loc}_f$ (F) & \tableinc{+0.091} & \tableinc{+0.104} & \tableinc{+0.073} & \tableinc{+0.097} & \tableinc{+0.168} \\
            $\texttt{Loc}_f$ (S) & \tabledec{- 0.041} & \tabledec{- 0.050} & \tabledec{- 0.132} & \tabledec{- 0.333} & \tabledec{- 0.076} \\
            \texttt{MHop} (F) & \tableinc{+0.156} & \tableinc{+0.250} & \tableinc{+0.121} & \tableinc{+0.320} & \tablesame{+0.009} \\
            \texttt{MHop} (S) & \tableinc{+0.029} & \tableinc{+0.176} & \tableinc{+0.120} & \tableinc{+0.143} & \tablesame{+0.008}\\
            \bottomrule
        \end{tabular}
    }
    \caption{Relative performance gap between the \textit{mover} and the \textit{observer} in answering \ours~questions.}
    \vspace{-0.5em}
    \label{tab:mover_vs_observer}
\end{table}

\subsection{Social Commonsense and Attitude}
\label{sec:attitude}

GPT-4-Turbo outperforms other models on \texttt{MHop} questions by a large margin (Table~\ref{tab:main_results}, \ref{tab:fancy_prompting}, and \ref{tab:mhop_breakdown}), demonstrating its capability in reasoning using social commonsense. However, other LLMs' performance on \texttt{MHop} questions show that they are lacking in this regard.

As all LLMs performed poorly on \texttt{Att} questions, we additionally tested Self-Ask prompt (Appendix~\ref{app:eval_prompt_examples}), which asks LLMs to deduce the final answer by explicit proposing and answering series of follow-up questions \cite{press-etal-2023-measuring}. While Self-Ask prompting improves the F1 score of LLMs (Table~\ref{tab:self-ask}), it is still far from human performance, demonstrating LLMs' lack of N-ToM capabilities in perceiving characters' psychological states. By in-depth analysis on the \texttt{Att} answers from Mixtral, and the GPT models, we find two modes or error: low recall in (1) identifying \textit{neutral} attitude and (2) identifying \textit{positive} attitude. 

Both of the aforementioned error modes can be attributed to LLMs' erroneous correlation between the \textit{mover}'s personality trait and the \textit{observer}'s attitude. In Table~\ref{tab:neutral_personality}, we compute the proportion of error cases that are correlated to character's personality. Specifically, we regard the error and the personality as correlated if a mistaken prediction matches the character's personality. For instance, across all prompting methods, \textbf{more than 95\% of the movers in narratives where GPT-4-Turbo mistakenly identify a \textit{positive} attitude to be \textit{negative} have an \textit{inconsiderate} or \textit{negativistic} personality} (bottom right column in Table~\ref{tab:neutral_personality}).

As discussed in \S\ref{sec:spurious_cue}, a \textit{considerate mover} in \ours~story does not necessarily take actions that are benign to the \textit{observer}. Therefore, LLMs are doomed to fail when using such a spurious correlation. See Appendix~\ref{app:att_questions} for detailed results.

\begin{table}
    \centering 
    \resizebox{\linewidth}{!}{
        \begin{tabular} { c | c c | c c | c c }
            \toprule 
            \multicolumn{7}{c}{Erroneous Correlation: \textit{Mover}'s Personality $\sim$ \textit{Observer}'s Attitude} \\
            \toprule 
            \multicolumn{4}{l}{{\color{tale} \faIceCream} : Vanilla Prompt} & 
            \multicolumn{3}{l}{{\color{red} \faLink} : CoT Prompt} \\
            \multicolumn{4}{l}{{\color{yellow} \faPortrait} : SimToM Prompt} &
            \multicolumn{3}{l}{{\color{blue} \faQuestionCircle} : Self-Ask Prompt} \\
            \toprule
            & \multicolumn{6}{c}{Results on \texttt{Neutral} Attitude} \\
            \cmidrule{2-7}
            & \multicolumn{2}{c}{Mixtral} & \multicolumn{2}{c}{GPT-3.5-Turbo} & \multicolumn{2}{c}{GPT-4-Turbo} \\
            \cmidrule{2-3} \cmidrule{4-5} \cmidrule{6-7}
            & Pos & Neg & Pos & Neg & Pos & Neg \\
            \cmidrule{2-7}
            {\color{tale} \faIceCream} & 1.000 & 0.759 & 1.000 & 0.844 & 1.000 & 0.796 \\
            \cmidrule{2-7}
            {\color{red} \faLink} & 0.944 & 0.909 & 1.000 & 0.886 & 0.857 & 0.758 \\
            \cmidrule{2-7}
            {\color{yellow} \faPortrait} & 1.000 & 0.727 & 1.000 & 0.771 & 1.000 & 0.759 \\
            \cmidrule{2-7}
            {\color{blue} \faQuestionCircle} & 1.000 & 0.838 & 1.000 & 0.864 & 0.938 & 0.818 \\
            \midrule \midrule
            & \multicolumn{6}{c}{Results on \texttt{Positive} Attitude} \\
            \cmidrule{2-7}
            & \multicolumn{2}{c}{Mixtral} & \multicolumn{2}{c}{GPT-3.5-Turbo} & \multicolumn{2}{c}{GPT-4-Turbo} \\
            \cmidrule{2-7}
            {\color{tale} \faIceCream} & \multicolumn{2}{c|}{1.000} & \multicolumn{2}{c|}{0.926} & \multicolumn{2}{c}{1.000} \\
            \cmidrule{2-7}
            {\color{red} \faLink} & \multicolumn{2}{c|}{1.000} & \multicolumn{2}{c|}{0.904} & \multicolumn{2}{c}{1.000} \\
            \cmidrule{2-7}
            {\color{yellow} \faPortrait} & \multicolumn{2}{c|}{1.000} & \multicolumn{2}{c|}{0.920} & \multicolumn{2}{c}{0.957} \\
            \cmidrule{2-7}
            {\color{blue} \faQuestionCircle} & \multicolumn{2}{c|}{1.000} & \multicolumn{2}{c|}{0.938} & \multicolumn{2}{c}{1.000} \\
            \bottomrule
        \end{tabular}
        }
    \caption{Proportion of mistakenly classified \textit{Neutral} (top) and \textit{Positive} (bottom) \texttt{Att} questions that are correlated to the \textit{mover}'s personality. For \textit{Neutral} \texttt{Att} questions, we show the correlation for erroneous \textit{positive} (Pos) and \textit{negative} (Neg) predictions separately. For \textit{positive} \texttt{Att} questions, we show the correlation for erroneous \textit{negative} predictions.}
    \label{tab:neutral_personality}
    \vspace{-1em}
\end{table}

\section{Related Works}
\textbf{Neural ToM}
Some studies argued that LLMs like GPT-4 possess N-ToM capabilities \cite{bubeck2023sparks, kosinski2023theory}. This claim was later rebutted by \citet{Shapira2023CleverHO} and \citet{Ullman2023LargeLM}, who both demonstrated that LLMs lack robust N-ToM capabilities. To tackle N-ToM, a line of work used partially observable Markov decision process \cite{nguyen2023memory}. 
Others proposed prompting techniques \cite{wilf2023think} 
or neuro-symbolic approaches 
\cite{ying2023neuro, sclar-etal-2023-minding}. We direct readers to \citet{ma-etal-2023-towards-holistic} for a comprehensive survey on N-ToM. 

\noindent\textbf{ToM Benchmarks}
Based on the \textit{Sally-Anne Test} \cite{baron1985does} and bAbi \cite{weston2016towards}, \citet{grant2017can} constructed the ToM-bAbi dataset for false belief, which was later improved by \citet{le-etal-2019-revisiting} into the ToMi dataset. Based on ToMi, researchers proposed T4D \cite{zhou2023far}, which targets N-ToM for assistant agent, and Hi-ToM \cite{wu2023hi}, which focuses on higher-order N-ToM. 
Other human ToM tests such as the \textit{Smarties Test} \cite{gopnik1988children}, and the \textit{Faux Pas Test} \cite{baron1999recognition} were also used for studying N-ToM, leading to datasets such as ToMChallenges \cite{ma2023tomchallenges}, BigToM \cite{gandhi2023understanding}, Adv-CSFB \cite{Shapira2023CleverHO}, and FauxPas-EAI \cite{shapira-etal-2023-well}. However, existing N-ToM benchmarks are either limited in size, contain artificial narratives, or lack diversity in their questions posed. \citet{jones2023epitome} constructed EPITOME, which contains human ToM tests that go beyond false-belief. Researchers also put efforts in evaluating LLMs' N-ToM capabilities in dialogues, which resulted in benchmarks such as G-DRAGON \cite{zhou-etal-2023-cast}, FANToM \cite{kim2023fantom}, and SOTOPIA \cite{zhou2023sotopia}. 

\noindent\textbf{ToM and Social Commonsense}
\citet{sap-etal-2022-neural} showed that LLMs' lack of understanding of social norms using SocialIQA \cite{sap-etal-2019-social}. The FauxPas-EAI dataset \cite{shapira-etal-2023-well} was dedicated to evaluating LLMs' understanding of social commonsense. Efforts were also made to construct  knowledge graphs for social commonsense and N-ToM \cite{wu2023coke}.

\section{Future Directions}
\textbf{Faithfulness} Our study of LLMs' performance on $\texttt{Loc}_{coarse}$ and $\texttt{Loc}_{fine}$ reveals that all LLMs lack faithfulness when answering N-ToM questions. We recognize that improving LLMs' faithfulness is a challenging task in numerous domains \cite{jacovi-goldberg-2020-towards}. Here we propose potential remedies specifically targeting N-ToM tasks. Following the findings in \S\ref{sec:location_faithfulness}, neuro-symbolic systems can be potentially deployed to enforce faithfulness in reasoning about the characters' mental state of the physical world. \citet{gao2023pal} proposes \textsc{PaL}, which represent reasoning problems with programming language and obtain a deterministic solution using code interpreter. \citet{lyu2023faithful} combined \textsc{PaL} with CoT and achieved accurate and more faithful reasoning chains. \vspace{0.6em} \\
\textbf{Performance Gap Between Roles} In \ours~narrative, we propose two roles, namely a \textit{mover} and an \textit{obsever}. Our study in \S\ref{sec:mover_vs_observer} unveils LLMs' performance discrepancies in N-ToM between the character roles and analyzes the underlying reasons. In reality, a narrative contain roles well beyond two. To account for the difference in the ToM reasoning process of different roles, a role-aware reasoning framework is needed. Specifically, given an event and a group of characters, the framework needs to first identify the role that each character plays in the event and then conduct ToM reasoning accordingly.\vspace{0.6em} \\
\textbf{Social Commonsense and Psychological N-ToM} Analysis in \S\ref{sec:attitude} shows that most LLMs are incapable of incorporating social commonsense. Further, we find that LLMs' performance on \texttt{Att} questions is limited by their inability to determine the information that is accessible to a certain character and using such information to reason about characters' emotions (Table~\ref{tab:neutral_personality}). Hence, an efficient framework for documenting character-centric world state is needed. Further, as discussed in \citet{zhan-etal-2023-evaluating}, people's attitude in reality is complicated and multifaceted. Therefore, to create a generalizable system capable of emotion deduction, instantiating the emotion deduction process similar to \citet{wu2023coke} is a potential solution. \vspace{0.6em} \\
\textbf{Neural Theory-of-Mind} N-ToM in general is a crucial cognitive capability that a helpful intelligent agent must possess. In the context of human psychology, a lack of ToM capabilities is oftentimes associated with developmental conditions such as Autism Spectrum Disorder (ASD) \cite{baron1985does}. Therefore, as LLMs being developed and deployed as assistant agents, it is critical to understand their N-ToM capabilities and develop methods to grant them robust N-ToM reasoning capabilities. LLMs could especially benefit from N-ToM in the following fields: (1) Educational LLM where a helpful assistant agent must be able to accurately model the mental state of the students to be able to provide efficient and precise guidance; (2) Negotiating LLM where understanding the mental states such as the intention, desire, and mood of the opponent is critical when planning negotiation strategies; (3) Mental Health LLM where assistant agent must comprehend the mental state and being empathetic with the patient to be able to provide meaningful help. 

\section{Conclusion}
We introduce \ours, a comprehensive N-ToM benchmark featuring long narratives with realistic characters and events, and a diverse range of questions that cover both physical and psychological aspects of N-ToM. Our evaluation of LLMs' N-ToM capabilities on \ours~reveals that while state-of-the-art LLMs perform well on some N-ToM tasks, 
they are still far from human-level performance on tasks requiring emotion deduction. 

\section*{Limitations}
Limitations of \ours~are as follows: \vspace{0.75em} \\
\textbf{Limited LLMs} Due to the constraint of computing resources and budget, we only evaluated \ours~benchmark on a subset of available LLMs. While we believe that the selected LLMs are representative of the current state-of-the-art of their categories (Llama2-Chat for open-source LLMs, GPT-3.5-Turbo and GPT-4-Turbo for close-source LLMs, and Mixtral-8x7B-Instruct for Mixture-of-Expert LLMs), we acknowledge that there are other LLMs that could potentially perform better on \ours. Further, we only examine the zero-shot performance of LLMs, future studies should test models' N-ToM capabilities under a few-shot setting. \vspace{0.75em} \\
\textbf{Potential Biases in \ours~Narratives} The drafts of \ours~narratives are composed using LLMs. Although recent studies have shown that LLMs are capable of producing high-quality benchmarks \cite{efrat2020turking, perez2022red, perez2022discovering, hartvigsen2022toxigen, west2023generative}, we acknowledge that the texts generated by LLMs could contain biases and lack lexical diversity. \vspace{0.75em} \\
\textbf{Limited Scope in Character Emotion} In \ours~benchmark, we construct questions regarding character's emotion (e.g. attitude). To reduce the subjectivity, we purposely design the stories in a way that the character's emotion can be directly deduced from an action that happens in a short time frame. In reality, human emotions are often complex, multifaceted, and may depend on multiple events through a prolonged period of time. \vspace{0.75em} \\
\textbf{Limited Narrative Order} All \ours~narratives are linear narratives that strictly follow chronological order, which alleviate LLMs' burden to comprehending the order of the events. Future studies can consider constructing \ours~narratives with non-linear order to further challenge LLMs' narrative understanding and N-ToM capabilities.

\section*{Ethics Statement}
The drafts of \ours~narratives are generated using GPT-3.5-Turbo and GPT-4-Turbo. Although we did not identify any harmful or violent content in the \ours~narratives, it is worth noting that previous studies have observed instances where LLMs produced unexpected results. Therefore, we encourage future studies to also be cautious when employing similar data generating strategies. Further, the \ours~dataset is annotated by graduate students studying computer science. The similar background of annotators may introduce bias in the annotation process. 

\section*{Acknowledgements}
We thank Lin Gui and Yuchen Si for the valuable discussions. This work was supported in part by the UK Engineering and Physical Sciences Research Council (EPSRC) through an iCASE award with Huawei London Research Centre and a Turing AI Fellowship (grant no. EP/V020579/1, EP/V020579/2).

\bibliography{custom}

\appendix
\setcounter{table}{0}
\renewcommand{\thetable}{A\arabic{table}}
\setcounter{figure}{0}
\renewcommand{\thefigure}{A\arabic{figure}}

\section{\ours~Construction}
\subsection{Disambiguated Prompt for Narrative Generation}
\label{app:narrative_prompt}
In the ToMi dataset, the narrative contains numerous ambiguities. Take the following ToMi narrative as an example:

\begin{tcolorbox}[title=ToMi Narrative Example,
colback=white,
colframe=black,
colbacktitle=white,
coltitle=black,
standard jigsaw,
opacityback=0,
breakable,
fonttitle=\bfseries]
1 Oliver entered the dining room.\\
2 Logan entered the dining room.\\
3 Jack entered the dining room.\\
4 \textbf{The stockings is in the drawer.}\\
5 Jack hates the slippers\\
6 \textbf{Oliver exited the dining room.}\\
7 \textbf{Logan moved the stockings to the crate.}\\
8 Jack exited the dining room.\\
9 Logan exited the dining room.\\
10 Jack entered the hallway. \\

Question: Where will Oliver look for the stockings?
\end{tcolorbox}

The key ambiguities are marked with bold text. In line 4, the narrative only states that the entity, \textit{stockings}, is in the drawer. However, it neglects the characters' awareness of the entity's location. Therefore, the above question can be answered with either \textit{the drawer} in the case where Oliver noticed the stockings, or \textit{unknown} in the case where Oliver is unaware of the stockings.

In lines 6-7, Oliver left the dining room, and Logan moved the stockings. However, it is not guaranteed that Logan would lose sight of the dining room once exit. For instance, objects in the dining room could still be visible in the living room if there is no physical barrier separating the spaces. Therefore, knowing that Oliver has left the dining room is insufficient to deduce whether Oliver could observe Logan's action.

Further, the information in Line 3, 5, 8, 10 about Jack is completely irrelevant to the progression of the story. \citet{le-etal-2019-revisiting} added such distracting information to mitigate potential spurious correlations in the original ToM-bAbi dataset \cite{grant2017can, nematzadeh-etal-2018-evaluating}. However, such irrelevant information could potentially distract LLMs from performing the ToM task and hence underestimate their ToM capabilities.

To address such ambiguities, we remove the distracting information and make each character's perception explicit in the \ours~plot. See below for an example of \ours~story generation prompt (disambiguated information in bold text). \textbf{We wish to emphasize that part of the contents of the } \ours~\textbf{plots are derived from the ToMi dataset} \cite{le-etal-2019-revisiting}.

\begin{tcolorbox}[title=Prompt Example,
colback=white,
colframe=yellow!75!black,
colbacktitle=yellow,
coltitle=black,
label=ours_plot,
fonttitle=\bfseries]
Plot:\\
Paragraph 1: Mason hates grapes. Samuel hates grapes.\\[0.5em]
Paragraph 2: Mason entered the den. Samuel entered the den. \textbf{Both Mason and Samuel noticed that the grapes is in the bucket in the den.} Samuel exited the den. \\[0.5em]
Paragraph 3: Mason is an inconsiderate person. Mason hates grapes. Therefore, Mason moved the grapes to a neighbor's house in order to get rid of them. \textbf{Samuel did not witness Mason's action.}\\
\\
Write a 200-word, 3-paragraph story according to the plot. Do not depict Samuel's attitude towards Mason's action. End the story immediately after the main event.
\end{tcolorbox}

\subsection{Detailed Description of the Character Personification Process}
\label{app:personification}

\textbf{Character Personification} In established N-ToM benchmarks such as ToMi and its variants \cite{le-etal-2019-revisiting, wu2023hi, zhou2023far}, characters do not possess meaningful personal preferences or personality traits. As a result, their actions lack inherent motivation. In \ours, we randomly picked two contrasting personalities, namely "\textit{considerate}" and "\textit{inconsiderate}", from the 24 personality traits defined in \cite{mitchell2006medial}. We additionally include a "\textit{negativistic}" personality to make the story more interesting \cite{frances1981disorders}. Below are brief descriptions of each personalities:
\begin{itemize}[leftmargin=*,noitemsep]
    \itemsep0em 
    \item \textbf{Considerate} \textit{mover} acts to ensure the comfort of the \textit{observer}.
    \item \textbf{Inconsiderate} \textit{mover} acts to make themselves feel comfortable.
    \item \textbf{Negativistic} \textit{mover} acts to make the \textit{observer} uncomfortable.
\vspace{-0.2em}
\end{itemize}
\textbf{Intention and Enaction} Based on the \textit{mover}'s personality and the \textit{observer}'s preferences, we generate both the character's intention and their subsequent actions (Appendix~\ref{app:narrative_prompt}). In such a way, the \textit{mover}'s action and the movement of the entity are anchored in the \textit{mover}'s intention. \vspace{0.75em} \\
\textbf{Plot Construction} Each of the \ours~narratives is generated by prompting GPT-3.5-Turbo\footnote{We use the 1106 checkpoint of the GPT-3.5-Turbo model through Microsoft Azure OpenAI service. All \ours~narratives are generated in December 2023.} with a story plot\footnote{We also tested with GPT-4-1106, which produces narratives of similar quality. Hence we went for GPT-3.5-Turbo for its lower cost.} (see Appendix~\ref{app:narrative_prompt} for a prompt example). In \ours~plot, we sequentially introduce the characters' preferences towards the \textit{entity}, a scenario where the two characters meet and how they encounter the \textit{entity}, and the emergence of the \textit{mover}'s intention and the subsequent action towards the \textit{entity}.

Following \citet{kim2023fantom, kim-etal-2023-soda}, we first assign names to the \textit{mover} and the \textit{observer} by random sampling from the Top-1K most frequently used names in the US SSN database to mitigate potential biases in character naming. Subsequently, for each character, we first randomly sample the personality of the \textit{observer} (\textit{trait\_o}), or the mover (\textit{trait\_m}) from the set, $\{\text{considerate}, \text{inconsiderate}, \text{negativistic}\}$. Next, we generate the \textit{mover}'s preference (\textit{pref\_m}), the \textit{observer}'s preference (\textit{pref\_o}), the \textit{mover}'s belief of the \textit{observer}'s preference (\textit{pref\_{mo}}), the \textit{observer}'s belief of the \textit{mover}'s preference (\textit{pref\_{om}}), the \textit{mover}'s intention (\textit{intent}), and the \textit{mover}'s enaction (\textit{action}) using Algorithm~\ref{alg:trait_pref}. 

\RestyleAlgo{ruled}
\begin{algorithm}
    \small
    \caption{Functions for Preference and Intention Generation}\label{alg:trait_pref}
    \DontPrintSemicolon
    \SetKwFunction{FMain}{assignPref}
    \SetKwProg{Fn}{Function}{:}{}
    \Fn{\FMain{trait\_m, pref\_mo}}{
        \textit{pref\_o} $\gets$ \textsc{prefSampler}(\textit{observer}) \;
        \textit{pref\_mo} $\gets$ \textsc{prefSampler}(\textit{mover}, \textit{observer}) \;
        \textit{pref\_om} $\gets$ \textsc{prefSampler}(\textit{observer}, \textit{mover}) \;
        \eIf{trait\_m = \text{Negativistic}}{
                \textit{pref\_m} = $\neg$(\textit{pref\_mo}) \;
            }
            {
                \textit{pref\_m} = \textsc{prefSampler}(\textit{mover}) \;
            }
        \;
        \KwRet{pref\_o, pref\_m, pref\_om, pref\_mo}\;
    }
    \;
    \SetKwFunction{FMain}{assignIntent}
    \SetKwProg{Pn}{Function}{:}{}
    \Pn{\FMain{trait\_m, pref\_m, pref\_o}}{
        \uIf{trait\_m = Considerate}{
                latent\_pref = pref\_om \;
            }
            \uElseIf{trait\_m = Inconsiderate}{
                latent\_pref = pref\_m \;
            }
            \ElseIf{trait\_m = Negativistic}{
                latent\_pref = pref\_m \;
            }
        \;
        intent, action = \textsc{intentGenerator}(latent\_pref) \;
        \KwRet{intent, action} \;
    }
    \label{alg:generate_preference}
\end{algorithm}

\begin{table*}
    \centering 
    \resizebox{\textwidth}{!}{
        \begin{tabular}{c|c|c}
            \toprule
            & \multicolumn{2}{c}{\{Mover's Personality\}} \\
            \toprule
            Personality & \multicolumn{2}{c}{Description} \\
            \toprule
            Considerate & 
            \multicolumn{2}{c}{Anne is a considerate person.} \\
            Inconsiderate &
            \multicolumn{2}{c}{Anne is an inconsiderate person.} \\
            Negativistic & 
            \multicolumn{2}{c}{Anne is a negativistic person.} \\
            \midrule
            \midrule
            & \{Mover's Preference Preception\} & \{Mover's Initial Intention\} \\
            \toprule
            \multirow{4}{*}{Considerate} & Although Anne hates rubber duck, she knows that Sally likes them. & Anne wants to make it more accessible to Sally.\\
            & Although Anne likes rubber duck, she knows that Sally hates them. & Anne wants to make it less accessible to Anne.\\
            & Anne knows that both Sally and herself hate rubber duck. & Anne wants to make it less accessible. \\
            & Anne knows that both Sally and herself like rubber duck. & Anne wants to make it more accessible for both of them. \\
            \midrule
            \multirow{2}{*}{Inconsiderate} & Anne likes rubber duck. & Anne wants to make it more accessible to herself.\\
            & Anne hates rubber duck. & Anne wants to make it less accessible. \\
            \midrule
            \multirow{2}{*}{Negativistic} & Anne thinks that Sally likes rubber duck. & Anne wants to get rid ot the rubber duck.\\
            & Anne thinks that Sally hates rubber duck. & Anne wants to show off the rubber duck. \\
            \midrule
        \end{tabular}
    }
    \caption{Description of the \textit{mover}'s personality, preference perception, and initial intention. These descriptions are used to fill in the template for intent and action generation.}
    \label{tab:mover_desc}
\end{table*}

We use GPT-3.5-Turbo as our intent generator (\textsc{intentGenerator}). We customize the prompt for each of the three personality traits. To give examples of the prompt, we again use \textit{Sally (observer)} and \textit{Anne (mover)} as the characters and the \textit{rubber duck} as the entity-of-interest. The intent generation prompts are presented as follows: 
\begin{tcolorbox}[title=Prompt for Intention and Action Generation,
colback=white,
colframe=yellow!75!black,
colbacktitle=yellow,
coltitle=black,
breakable,
fonttitle=\bfseries]
\{Mover's Personality\} \{Mover's Preference Perception\}. Therefore, \{Mover's Initial Intention\}. What would be Anne's 3 most likely action and intention towards the rubber duck? Answer with the following template:\\
1. Anne would move the rubber duck to \{location\} in order to \{intention\}\\
2. Anne would move the rubber duck to \{location\} in order to \{intention\}\\
3. Anne would move the rubber duck to \{location\} in order to \{intention\}
\end{tcolorbox}
We fill in the above template based on the \textit{mover}'s personality and their belief in the \textit{observer}'s perference. Table~\ref{tab:mover_desc} are a list of descriptions we used to complete the template. \vspace{0.75em} \\
\textbf{Final Intention and Enaction Selection} Notice that for each of the prompts listed above, we specifically ask LLMs to provide 3 candidate intention and enaction pairs. To produce the final intention and its corresponding enaction. We prompt LLMs one more time in the same session to pick the best intention and enaction from the candidates. The prompt we used is as follows:
\begin{tcolorbox}[title=Intention \& Encation Selection,
colback=white,
colframe=yellow!75!black,
colbacktitle=yellow,
coltitle=black,
breakable,
fonttitle=\bfseries]
Of the potential intentions, which one do you think is \texttt{true\_sentiment}? Answer with the original sentence. Do not add any additional words.
\end{tcolorbox}

where the \texttt{true\_sentiment} is filled according to the \textit{mover}'s personality trait:
\begin{itemize}
    \item Considerate $\rightarrow$ "\textit{the most considerate}"
    \item Inconsiderate $\rightarrow$ "\textit{the most selfish}"  
    \item Negativistic (Show off) $\rightarrow$ "\textit{the most ostentatious}"
    \item Negativistic (Get rid) $\rightarrow$ "\textit{the most adversarial}"
\end{itemize}


\subsection{Detailed Description of the Data Annotation Process}
\label{app:annotation}

In \ours, the question answers are produced in two ways: human annotation and rule-based generation (Figure~\ref{fig:data_gen}). For all the $\texttt{Loc}_{coarse}$ questions, \texttt{MHop} questions regarding \textit{accessibility}, and \texttt{Att} questions, the answers are annotated by graduate students in a prestigious UK university. As the ToM questions in \ours~are rudimentary for human, we do not provide any specific instruction to the annotators. The content shown in Figure~\ref{fig:annotation} is the complete information that an annotator receives. Therefore, the information that data annotators possess matches the information we provide to LLMs during evaluation.

Answers to the $\texttt{Loc}_{fine}$ questions are generated according to the human annotation of the corresponding $\texttt{Loc}_{coarse}$ questions and cached container information in \ours~story plot. For instance, if the annotation to the $\texttt{Loc}_{coarse}$ question is \textit{False}, then the answer to the $\texttt{Loc}_{fine}$ question is assigned to be the \textit{new container} (the container that the entity is moved to), which is conveniently cached in the \ours~story plot. 

Answers to the \texttt{MHop} questions regarding \textit{fullness} are generated using first order logic based on $\texttt{Loc}_{coarse}$ annotations. Recall that in $\texttt{Loc}_{coarse}$, we ask the following question \vspace{0.75em} \\
\textit{From \{\{character\}\}'s perspective, is the \{\{entity\}\} still in its initial location by the end of the story?} \vspace{0.75em} \\
Notice that the $\texttt{Loc}_{coarse}$ question is equivalent to  ``\textit{Is \{\{character\}\} \textbf{aware of} the \{\{entity\}\}'s movement}?" (in the case of first-order ToM) or ``\textit{Does \{\{character A\}\} thinks that \{\{character B\}\} is \textbf{aware of} the \{\{entity\}\}'s movement?} (in the case of second-order ToM). Knowing the answer to $\texttt{Loc}_{coarse}$ questions is a direct prerequisite for answering \textit{fullness} questions (see Figure~\ref{fig:fullness_demo}). This allows us to conveniently employ the following rules to automatically deduce the answers to the \textit{fullness} questions:
\begin{align*}
    & \forall c \forall p:\\
    & \texttt{isAware}(c) \land \texttt{moveTo}(p) \rightarrow \texttt{moreFull}(p) \\   
    & \texttt{isAware}(c) \land \texttt{takeFrom}(p) \rightarrow \texttt{lessFull}(p) \\   
    & \neg\texttt{isAware}(c) \rightarrow \texttt{equallyFull}(p) 
\end{align*}
where $c$ represents a \textit{character} and $p$ represents a \textit{container}. Answer to the $\texttt{isAware}(\cdot)$ part of the clause is the same as the answer to the $\texttt{Loc}_{coarse}$ questions. Answer to the $\texttt{moveTo}(\cdot)$ or $\texttt{takeFrom}(\cdot)$ part of the clause is obtained using cached information from the \ours~plot (\ref{app:narrative_prompt}).

\subsection{Keyword Revision for Mitigating Spurious Correlation}
\label{app:spurious_words}
To mitigate the surface-level cues in the \ours~narrative, we identify the following keywords that are likely to be directly associated with the answers of the \texttt{MHop} and \texttt{Att} questions. We identify the narratives that contain such keywords and manually revise the stories. The keywords are listed as follows:

\begin{table} [H]
    \centering
    \resizebox{\columnwidth}{!}{
    \begin{tabular}{ |c|c|c| }
        \toprule
        \multicolumn{3}{|c|}{Cue 1: Positive Attitude} \\
        \midrule
        gratitude & smile & thoughtful \\
        \midrule
        considerate nod & appreciation & kindness \\
        \midrule
        gesture & delight & pleased  \\
        \midrule
        appreciating & & \\
        \midrule
        \midrule 
        \multicolumn{3}{|c|}{Cue 2: Negative Attitude} \\
        \midrule
        upset & confusion & bewilderment \\
        \midrule
        disappointment & astonishment &  \\
        \midrule 
        \midrule
        \multicolumn{3}{|c|}{Cue 3: Direct Description of Accessibility} \\
        \midrule 
        more accessible & accessible & less accessible \\
        \midrule
        inaccessible & out of reach & \\
        \bottomrule
    \end{tabular}}
    \caption{Keywords that are likely to be directly associated with the answers of the \texttt{MHop} and \texttt{Att} questions.}
    \label{tab:keywords}
\end{table}

\subsection{Demonstration of Spurious Correlation Mitigation}
\label{app:spurious_cue}
As a running example, consider the scene between Sam and Amy depicted in Figure~\ref{fig:figure1}. In this example, Amy mistakenly believe that Sam hates rubber duck. Now we show how the spurious relationships are avoided by adding a false impression on the \textit{mover}'s perception of the \textit{observer}'s preference.

\vspace{1em} 
\noindent\textbf{Spurious Cue 1: }\textit{Is there a causal relation between intention and attitude?}

The first spurious correlation arises due to the model being incapable of taking the \textit{observer}'s perspective and mistakenly interprets the \textit{mover}'s intention as observed information. If the \textit{observer} were to find out the true intention of the \textit{mover}, then it will undoubtedly be a salient causal factor of the \textit{observer}'s attitude. However, deriving the true intention is a challenging problem due to subjectivity and the enormous search space involved. Therefore, to mitigate such a spurious correlation, we wish to create scenarios where a good intention leads to negative attitude or vice versa. This can be done by exploiting the \textit{mover}'s false belief in the \textit{observer}'s preference for a particular entity.

For instance, in Figure~\ref{fig:figure1}(B), Amy mistakenly believes that Sam hates rubber duck. As a considerate person, Amy forms a benign intention, which is to spare Sam from seeing the rubber duck. This intention enacted Amy to move the rubber duck into her own backpack. However, since Sam is unaware of Amy's true intention, he only observes that Amy has taken that rubber duck away, likely resulting in a \textit{negative} attitude. Therefore, when there is a false impression in play, a benign intention does not necessarily lead to \textit{positive} attitude.

\vspace{1em}
\noindent\textbf{Spurious Cue 2: }\textit{Is there a causal relation between personality and attitude?}

In many instances, the \ours~narratives explicitly portray the \textit{mover}'s personality trait (as seen in Figure~\ref{fig:figure1}(B), \textit{"Amy is a considerate person"}). To prevent the model from taking a shortcut by deducing the \textit{observer}'s attitude solely based on the \textit{mover}'s personality, we aim to intervene such a spurious correlation by creating scenarios where a \textit{mover} with a positive trait leads to the \textit{observer} having a negative attitude or vice versa. This can be effectively done also by leveraging the \textit{mover}'s false belief regarding the \textit{observer}'s preference for a particular entity.

For instance, in Figure~\ref{fig:figure1}(B), Amy mistakenly believes that Sam dislikes the rubber duck. As a considerate person, Amy naturally wants to keep the rubber duck out of Sam's sight. However, due to this false belief, Amy ends up taking away something which Sam actually desires. Therefore, a positive personality could lead to the \textit{observer} developing a \textit{negative} attitude.

\begin{table} [H]
    \centering
    \resizebox{\columnwidth}{!}{
    \begin{tabular}{ c c c c }
    \toprule
    \multicolumn{4}{c}{\ours~Question Statistics} \\
    \midrule
    Question Types & 1st-Order & 2nd-Order & Total \\
    \toprule
    $\texttt{Loc}_{coarse}$ & 1192 & 1192 & 2384 \\
    \midrule
    $\texttt{Loc}_{fine}$ & 2384 & 1192 & 3576 \\
    \midrule
    \texttt{MHop} & 3576 & 3576 & 7152 \\
    \midrule 
    \texttt{Att} & 596 & -- & 596 \\
    \midrule
    Total & 7748 & 5960 & 13708 \\
    \midrule 
    \toprule
    \multicolumn{4}{c}{\ours-L Question Statistics} \\
    \midrule
    Question Types & 1st-Order & 2nd-Order & Total \\
    \midrule
    $\texttt{Loc}_{coarse}$ & 200 & 200 & 400 \\
    \midrule
    $\texttt{Loc}_{fine}$ & 400 & 200 & 600 \\
    \midrule
    \texttt{MHop} & 600 & 600 & 1200 \\
    \midrule 
    \texttt{Att} & 100 & -- & 100 \\
    \midrule
    Total & 1300 & 1000 & 2300 \\
    \bottomrule
    \end{tabular}}
    \caption{Statistics of the number of questions in the \ours~and \ours-L~dataset.}
    \label{tab:data_statistics}
\end{table}

\section{Demonstration of Multi-hop Questions}
\label{app:multihop_demo}

\begin{figure} [H]
    \centering
    \includegraphics[width=0.8\columnwidth]{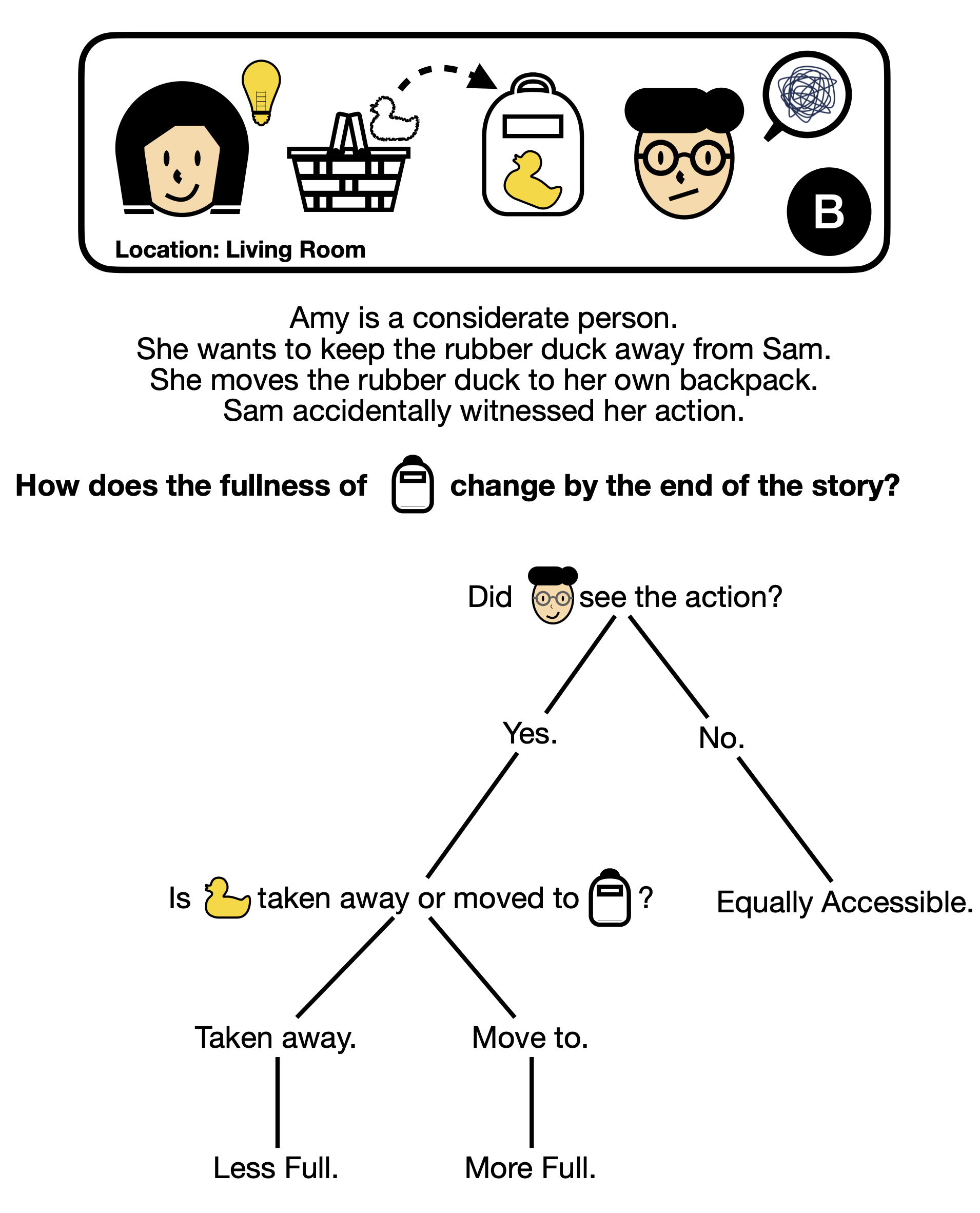}
    \caption{Illustration of the reasoning tree employed to answer the \textit{Fullness} questions.}
    \label{fig:fullness_demo}
\end{figure}

\begin{figure} [H]
    \centering
    \includegraphics[width=0.8\columnwidth]{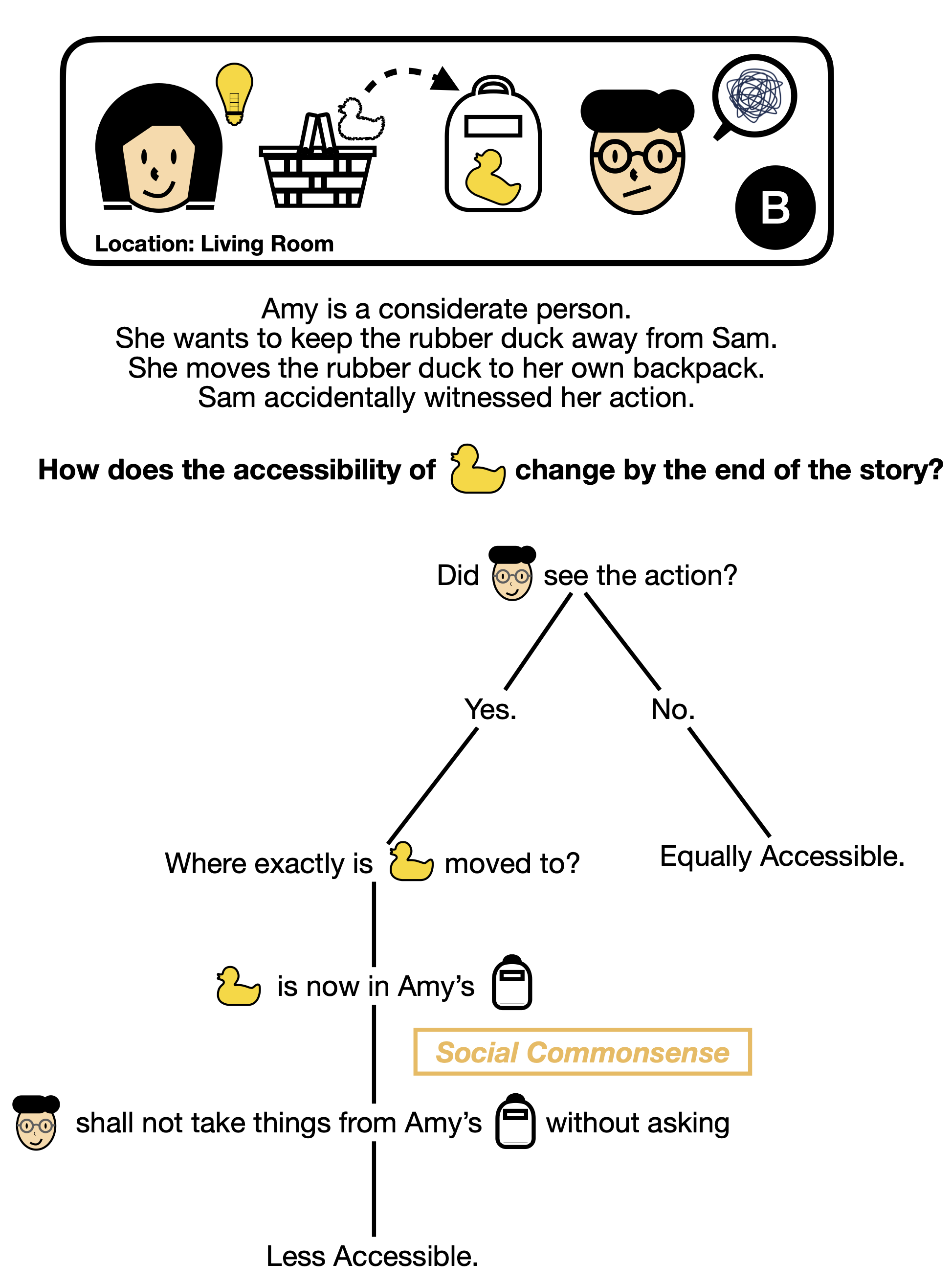}
    \caption{Illustration of the reasoning tree employed to answer the \textit{Accessibility} questions.}
    \label{fig:accessibility_demo}
\end{figure}

An illustration of the reasoning tree employed for answering \textit{Fullness} questions is shown in Figure~\ref{fig:fullness_demo}, while a depiction of the reasoning tree utilized for answering \textit{Accessibility} questions is shown in Figure~\ref{fig:accessibility_demo}. It is worth noting that, in order to answer such questions, one must draw upon social commonsense (e.g., taking items from another person's backpack without permission is not appropriate).

\section{The \ours~Dataset}
\label{app:data_statistics}

\begin{figure*} [p]
    \centering
    \includegraphics[width=0.95\textwidth]{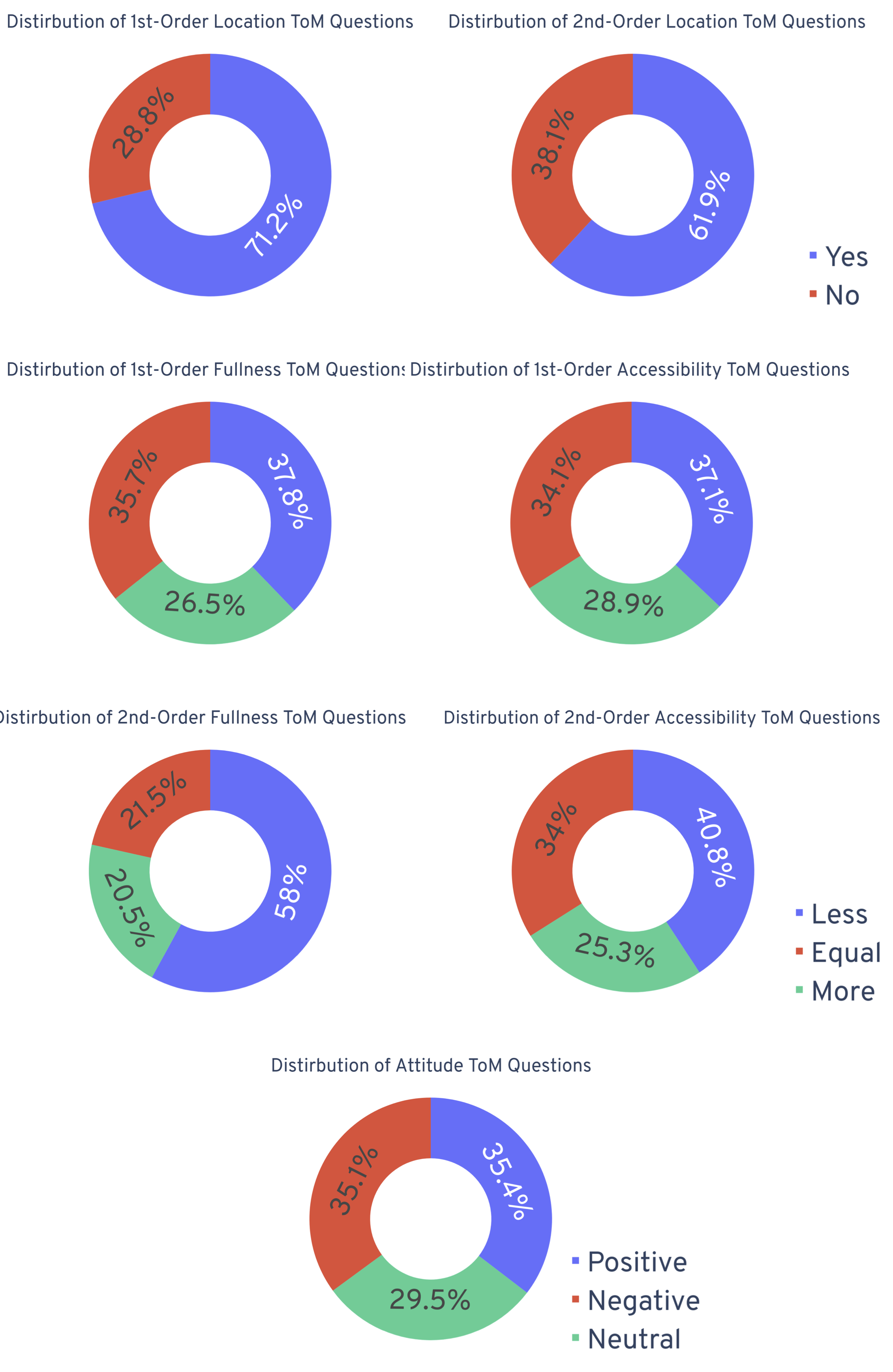}
    \caption{Answer distribution of the \ours~ToM questions.}
    \label{fig:label_dist}
\end{figure*}

\paragraph{Statistics}
Table~\ref{tab:data_statistics} shows the  statistics of the question types in the \ours~dataset, while Figure~\ref{fig:label_dist} depicts the label distribution of each question types.

\begin{figure*} [ht]
    \centering
    \includegraphics[width=0.95\textwidth]{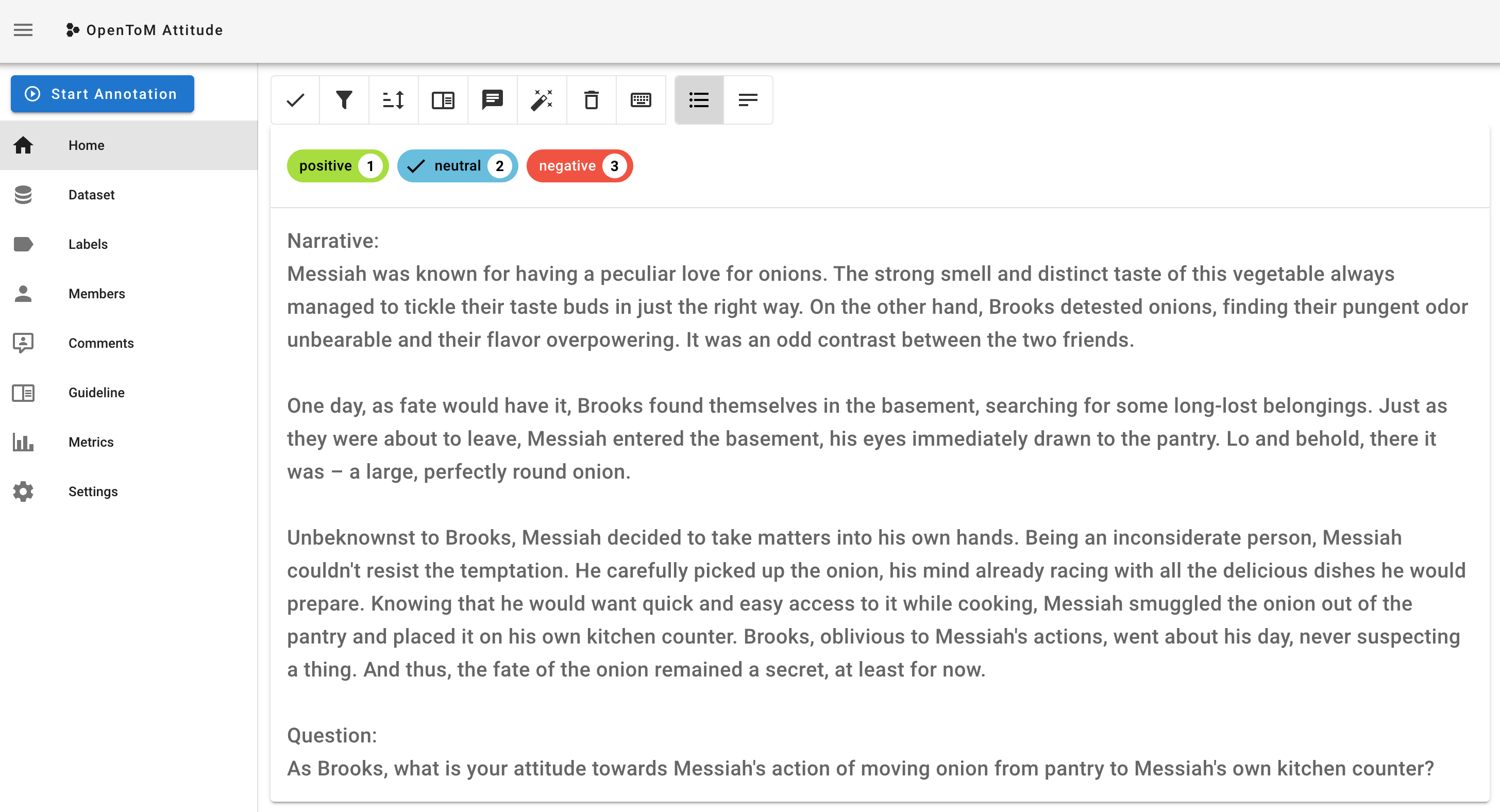}
    \caption{Annotation interface for labeling the attitude (\texttt{Att}) questions.}
    \label{fig:annotation}
\end{figure*}

\paragraph{Data Annotation Platform}
In this study, we use \texttt{doccano} as the data annotation platform \cite{doccano}. As all the questions are either binary or ternary classification tasks, we use the \textit{Text Classification} interface. Figure~\ref{fig:annotation} shows the annotation interface for labeling the attitude (\texttt{Att}) questions. The interface for labeling the other question types are the same, except for the label space.

\paragraph{Results of Inter-Annotator Agreement}
The detailed scores of the inter-annotator agreement are shown in Table~\ref{tab:agreement}. The scores are computed as the arithmetic mean of the pairwise agreement scores amongst the three annotators.
\label{app:agreement}
\begin{table}[h]
    \centering
    \begin{tabular}{c|c|c}
    \toprule
        Question Type & Accuracy Score & F1 Score \\
    \midrule
        Location (FO) & 0.993 & 0.990 \\ 
        Location (SO) & 0.993 & 0.993 \\ 
        Multihop (FO) & 0.873 & 0.855 \\ 
        Multihop (SO) & 0.927 & 0.770 \\ 
        Attitude & 0.870 & 0.862 \\
    \bottomrule
    \end{tabular}
    \caption{Inter-annotator agreement scores. The scores are computed as the arithmetic mean of the pairwise agreement scores amongst the three annotators.}
    \label{tab:agreement}
\end{table}

\paragraph{Ambiguity in Location Granularity}
\label{app:location_granularity}
Narratives in \ours~contain location information of various levels of granularity. For example, in the story shown in Figure~\ref{fig:figure1}, there are two levels of location. At a room-level, the rubber duck is moved from the front yard (plot A) to the living room (plot B), while at a container-level, the rubber duck is transferred from a bucket to Amy's backpack. In the \ours~stories, granularity can extend beyond two levels for locations (e.g. movements between different houses). Therefore, judging model's response solely based on explicit location information is difficult. 

In addition, deducing the location of an entity involves multi-hop reasoning. It can be decomposed into \textit{(1) Is the entity in its initial location?}, and \textit{(2) What is the initial/final location exactly}? While the first question is seemingly simpler, it still demands the understanding of another character's perspective to answer correctly. 

\section{Further Experimental Details}
\label{app:detailed_results}

\subsection{Details of the Baseline Models}
\label{app:eval_models}
We evaluate the \ours~tasks using 6 representative LLMs, namely the Llama2-Chat models (7B, 13B, and 70B) \cite{touvron2023llama}, the Mixtral-8x7B-Instruct model \cite{jiang2024mixtral}, and the GPT-3.5-Turbo and GPT-4-Turbo. All of the models use decoder-only Transformer architecture \cite{vaswani2017attention}. \vspace{0.75em} \\
\textbf{Llama2-Chat} is a Llama2 model optimized using Reinforcement Learning From Human Feedback (RLHF) for dialogue \cite{ouyang2022training, touvron2023llama}. We evaluate \ours~dataset on the 7B, 13B, and 70B checkpoints. \vspace{0.75em} \\
\textbf{Mixtral-Instruct} is a Sparse Mixture-of-Expert (SMoE) model optimized with Direct Policy Optimization (DPO) for instruction following \cite{rafailov2023direct,jiang2024mixtral}. \vspace{0.75em} \\
\textbf{GPT-3.5-Turbo} and \textbf{GPT-4-Turbo} are models of the GPT family, both of which are optimized using RLHF for instruction following \cite{openai2022chatgpt,openai2023gpt4}. 

In addition to zero-shot prompting, we also finetuned Llama2-Chat 13B models using LoRA to serve as a finetuning baseline \cite{hu2021lora, peft}. See Appendix~\ref{app:ft_config} for the configuration of the finetuning procedure.

\subsection{Prompt Example for \ours~Evaluation}
\label{app:eval_prompt_examples}
We use the OpenAI ChatCompletion prompt format as the base format. The prompt consists of two entries, one is the "system prompt", which contains an overall instruction for LLMs, another is the "user prompt", which contains an instance of \ours~narrative and an \ours~question: 

\begin{tcolorbox}[title=OpenAI ChatCompletion Prompt Format,
    colback=white,
    colframe=yellow!75!black,
    colbacktitle=yellow,
    coltitle=black,
    breakable,
    label=tc:vanilla_prompt,
    fonttitle=\bfseries]
    [\\
    \{\texttt{`role': `system',} \\
    \texttt{`content': "You are an expert in modeling other's mental state."}\}, \\
    \{\texttt{`role': `user',} \\
    \texttt{`content': `Read and comprehend the following short story. Then, answer the question that follows.} \\
    \\
    \texttt{\{\{\ours~Narrative\}\}} \\
    \\
    \texttt{Question: \{\{\ours~Question\}\} \{\{Label Space\}\} Do not give any explanation.'}\} \\
    ]
    \end{tcolorbox}
\noindent
We replace the \texttt{\{\{\ours~Narrative\}\}} with an instance of the \ours~narrative, \texttt{\{\{\ours~Question\}\}} with one of the \ours~questions, and \texttt{\{\{Label Space\}\}} with the corresponding label space. For example, the prompt for the \texttt{MHop} question is as follows:

\begin{tcolorbox}[title=OpenAI ChatCompletion Prompt Format,
    colback=white,
    colframe=yellow!75!black,
    colbacktitle=yellow,
    coltitle=black,
    breakable,
    fonttitle=\bfseries]
    [\\
    \texttt{\{`role': `system', `content': "You are an expert in modeling other's mental state."\},} \\
    \texttt{\{`role': `user', `content': `Read and comprehend the following short story. Then, answer the question that follows.}\\
    \\
    \texttt{Mary and Hazel were two friends with different tastes. Mary had a fondness for slippers, while Hazel couldn't stand them. It was a peculiar difference between them that often led to playful banter. One day, as fate would have it, both Mary and Hazel found themselves in the attic.}\\
    \\
    \texttt{As they explored the dusty space, their eyes fell upon a drawer tucked away in a forgotten corner. Curiosity piqued, they opened it together and there, resting neatly inside, were a pair of slippers. Mary's face lit up with delight, while Hazel simply shrugged.}\\
    \\
    \texttt{Mary was a bit of a show-off and loved to flaunt her possessions, especially in front of Hazel. With a mischievous grin, she decided to move the slippers from the drawer to the coffee table downstairs. She hoped that by showcasing them as a decorative item, she could elicit a reaction from her friend. Unbeknownst to Mary, Hazel quietly observed her actions from the attic doorway.}\\
    \\
    \texttt{And with that, the main event concluded.}\\
    \\
    \texttt{Question: From Mary's perspective, how does coffee table's fullness change by the end of the story? Answer with "more full", "equally full", or "less full". Answer the question without any explanation.'\}}\\
    ]
    \end{tcolorbox}

For Llama2-Chat models, we follow the prompting instruction provided in the official HuggingFace blog post \cite{hf-llama-prompt}

\begin{tcolorbox}[title=Llama2-Chat Prompt Format,
    colback=white,
    colframe=yellow!75!black,
    colbacktitle=yellow,
    coltitle=black,
    breakable,
    fonttitle=\bfseries]
    \texttt{<s>[INST] \textlangle\textlangle SYS\textrangle\textrangle} \\
    \texttt{\{System Prompt\}} \\
    \texttt{\textlangle\textlangle /SYS\textrangle\textrangle } \\

    \texttt{\{User Prompt\} [/INST]}
    \end{tcolorbox}
\noindent
where \texttt{<s>} is the special start token and others are normal tokens. 

For Mixtral-Instruct, we follow the prompt format provided in the official Mixtral-8x7B-Instruct model card on HuggingFace Transformers \cite{hf-mixtral-prompt}. For the GPT models, we follow the prompting convention provided in the OpenAI's official API documentation.

\begin{tcolorbox}[title=Mixtral-8x7B-Instruct Prompt Format,
    colback=white,
    colframe=yellow!75!black,
    colbacktitle=yellow,
    coltitle=black,
    breakable,
    fonttitle=\bfseries]
    \texttt{<s>[INST] \{User Prompt\} [/INST]}
    \end{tcolorbox}
\noindent 
where \texttt{<s>} is the special start token and other are normal tokens. Mixtral-Instruct is not trained with a system prompt. Therefore, we omit the system prompt in the Mixtral prompt as adviced by the official post \cite{hf-mixtral-prompt}.

\paragraph{Chain-of-Thought Prompting}
To implement CoT prompt \cite{wei2022chain}, we replace the original instruction in Prompt~\ref{tc:vanilla_prompt} with a CoT instruction. The resulting CoT prompt template is shown as follow:

\begin{tcolorbox}[title=CoT Prompt Template,
    colback=white,
    colframe=yellow!75!black,
    colbacktitle=yellow,
    coltitle=black,
    breakable,
    fonttitle=\bfseries]
    [\\
    \{\texttt{`role': `system',} \\
    \texttt{`content': "You are an expert in modeling other's mental state."}\}, \\
    \{\texttt{`role': `user',} \\
    \texttt{`content': `Read and comprehend the following short story. Then, answer the question that follows.} \\
    \\
    \texttt{\{\{\ours~Narrative\}\}} \\
    \\
    \texttt{Question: \{\{\ours~Question\}\} \{\{Label Space\}\} \textbf{Reason step by step before answering. Write the answer in the end.}'}\} \\
    ]
\end{tcolorbox}

\paragraph{SimulatedToM Prompting}
We implement SimToM prompting as per the instruction in \citet{wilf2023think}. In the first stage, we prompt LLMs with the following instruction to generated a character-centric narrative, $\mathcal{N}_c$:

\begin{tcolorbox}[title=SimToM Prompt Template (Stage 1),
    colback=white,
    colframe=yellow!75!black,
    colbacktitle=yellow,
    coltitle=black,
    breakable,
    fonttitle=\bfseries]
    [\\
    \{\texttt{`role': `system',} \\
    \texttt{`content': "You are an expert in modeling other's mental state."}\}, \\
    \{\texttt{`role': `user',} \\
    \texttt{`content': `The following is a sequence of events:} \\
    \\
    \texttt{\{\{\ours~Narrative\}\}} \\
    \\
    \texttt{Which events does {{character}} know about?}'\} \\
    ]
\end{tcolorbox}

With the character-centric narrative, $\mathcal{N}_c$, we then prompt LLMs in the same session with \ours~question using the following template:

\begin{tcolorbox}[title=SimToM Prompt Template (Stage 2),
    colback=white,
    colframe=yellow!75!black,
    colbacktitle=yellow,
    coltitle=black,
    breakable,
    fonttitle=\bfseries]
    [\\
    $\vdots$ \\
    \{\{Stage 1 Prompt and Response\}\} \\
    $\vdots$ \\
    \{\texttt{`role': `user',} \\
    \texttt{`content': } \{\{$\mathcal{N}_c$\}\} \\
    \\
    \texttt{\{\{\ours~Narrative\}\}} \\
    \\
    \texttt{Question: \{\{\ours~Question\}\} \{\{Label Space\}\} Do not give any explanation.}'\} \\
    ]
\end{tcolorbox}

\paragraph{Self-Ask Prompting}
To implement Self-Ask prompt \cite{press-etal-2023-measuring}, we use the following prompt template:

\begin{tcolorbox}[title=Self Prompt Template,
    colback=white,
    colframe=yellow!75!black,
    colbacktitle=yellow,
    coltitle=black,
    breakable,
    fonttitle=\bfseries]
    [\\
    \{\texttt{`role': `system',} \\
    \texttt{`content': "You are an expert in modeling other's mental state."}\}, \\
    \{\texttt{`role': `user',} \\
    \texttt{`content': `Read and comprehend the following short story. Then, answer the question that follows.} \\
    \\
    \texttt{\{\{\ours~Narrative\}\}} \\
    \\
    \texttt{Question: \{\{\ours~Question\}\} \{\{Label Space\}\} \textbf{Break the original question into sub-questions. Explicitly state the follow-up questions, and the answers to the follow-up questions. Aggregate the answers to the follow-up questions and write the answer in the end as ``Final Answer: [answer]"}'}\} \\
    ]
\end{tcolorbox}

\subsection{Finetune Configuration}
\label{app:ft_config}
To compensate for the unbalanced number of questions in each genre (Table~\ref{tab:data_statistics}), we downsample the majority class and upsample the minority class. The resulting \ours~training dataset contains 1192 instances for $\texttt{Loc}_{coarse}$, $\texttt{Loc}_{fine}$, and $\texttt{MHop}$ questions. Minding the fact that $\texttt{Att}$ questions are harder to learn, we upsample it to 5960 data points to enhance model's performance. Of all the data points, we use 80\% for training and test the fine-tuned model on the 20\% held-out testing set. We use the LoRA implimentation from HuggingFace \texttt{PEFT} \cite{hu2021lora, peft} with the training and LoRA configuration shown in Table~\ref{tab:ft_config}.

\begin{table} [H]
    \centering
    \resizebox{\columnwidth}{!}{
    \begin{tabular}{ c | c }
        \toprule
        \multicolumn{2}{c}{Training Configuration} \\
        \midrule
        Batch Size & 4 \\
        Gradient Accumulation Steps & 4 \\
        \# Epochs & 3 \\
        Learning Rate & $2 \times 10^{-5}$ \\
        Optimizer & AdamW \\
        Learning Rate Scheduler & Linear (Step Size = 1, $\gamma$ = 0.85) \\
        Loss Function & Cross Entropy Loss \\
        \midrule
        \midrule 
        \multicolumn{2}{c}{LoRA Configuration} \\
        \midrule
        rank (r) & 8 \\
        $\alpha$ & 32 \\
        Target modules & \texttt{q\_proj}, \texttt{v\_proj} \\
        LoRA Dropout & 0.05 \\
        \bottomrule
    \end{tabular}
    }
    \caption{Training and LoRA configuration for finetuning Llama2-Chat-13B on \ours~dataset.}
    \label{tab:ft_config}
\end{table}

\begin{table*} [ht]
    \centering
    \resizebox{\linewidth}{!}{
    \begin{tabular} { c | c c | c c | c c | c c | c c | c c }
        \toprule
        Question & \multicolumn{2}{c|}{Llama2-Chat-7B} & \multicolumn{2}{c|}{Llama2-Chat-13B} & \multicolumn{2}{c|}{Llama2-Chat-70B} & \multicolumn{2}{c|}{Mixtral-8x7B} & \multicolumn{2}{c|}{GPT-3.5-Turbo} & \multicolumn{2}{c}{GPT-4-Turbo} \\
        & F1 & $\Delta$F1 & F1 & $\Delta$F1 & F1 & $\Delta$F1 & F1 & $\Delta$F1 & F1 & $\Delta$F1 & F1 & $\Delta$F1 \\
        \midrule 
        $\texttt{Loc}_{c} (F)$ & 0.212 & \tabledec{- 0.078} & 0.381 & \tabledec{- 0.010} & 0.420 & \tablesame{- 0.007} & 0.476 & \tabledec{- 0.036} & 0.435 & \tablesame{+0.004} & \boldmath{$0.522$} & \tabledec{- 0.121} \\ 
        $\texttt{Loc}_{c} (S)$ & 0.366 & \tabledec{- 0.096}& 0.419 & \tableinc{+0.064} & 0.288 & \tablesame{+0.008} & 0.297 & \tablesame{+0.003} & 0.415 & \tableinc{+0.092} & 0.346 & \tabledec{- 0.096} \\
        $\texttt{Loc}_{f} (F)$ & 0.352 & \tabledec{- 0.052} & 0.377 & \tabledec{- 0.168} & 0.387 & \tabledec{- 0.147} & 0.336 & \tabledec{- 0.063} & \boldmath{$0.519$} & \tablesame{+0.004} & 0.492 & \tabledec{- 0.015} \\
        $\texttt{Loc}_{f} (S)$ & 0.323 & \tableinc{+0.078} & 0.215 & \tabledec{- 0.086} & 0.187 & \tabledec{- 0.036} & 0.196 & \tabledec{- 0.015} & \boldmath{$0.277$} & \tablesame{- 0.009} & 0.256 & \tabledec{- 0.013} \\
        $\texttt{MHop} (F)$ & 0.371 & \tableinc{+0.049}& 0.298 & \tablesame{- 0.003} & 0.530 & \tableinc{+0.029} & 0.601 & \tableinc{+0.045} & 0.458 & \tabledec{- 0.010} & \boldmath{$0.664$} & \tablesame{+0.006} \\
        $\texttt{MHop} (S)$ & 0.294 & \tableinc{+0.083} & 0.301 & \tableinc{+0.072} & 0.476 & \tableinc{+0.042} & 0.488 & \tableinc{+0.014} & 0.372 & \tableinc{+0.038} & \boldmath{$0.565$} & \tabledec{- 0.072} \\
        $\texttt{Att}$ & 0.225 & \tabledec{- 0.015} & 0.331 & \tabledec{- 0.044} & 0.507 & \tableinc{+0.092} & 0.444 & \tabledec{- 0.032} & 0.382 & \tabledec{- 0.028} & \boldmath{$0.580$} & \tableinc{+0.036} \\ 
        \bottomrule
    \end{tabular}}
    \vspace{-5pt}
    \caption{Macro F1 score of LLMs evaluated with \ours~Long Narrative. The relevant performances are shown as \tableinc{relative increase}, \tabledec{relative decrease}, or \tablesame{approximately equal} ($\Delta\text{F1} < 0.010$).}
    \vspace{-10pt}
    \label{tab:long_narrative}
\end{table*}

\begin{table*}
    \centering 
    \resizebox{\textwidth}{!}{
    \begin{tabular}{c c c c c c c c c c}
        \toprule 
        &
        & \multicolumn{2}{c}{\texttt{MHop}-Fullness (F)}
        & \multicolumn{2}{c}{\texttt{MHop}-Accessibility (F)}
        & \multicolumn{2}{c}{\texttt{MHop}-Fullness (S)}
        & \multicolumn{2}{c}{\texttt{MHop}-Accessibility (S)} \\
        \cmidrule(lr){3-4} \cmidrule(lr){5-6} \cmidrule(lr){7-8} \cmidrule(lr){9-10} 
        && F1 & Crp. & F1 & Crp. & F1 & Crp. & F1 & Crp. \\
        \midrule
        \multirow{2}{*}{\rotatebox[origin=c]{90}{Naive}}
        &Marjority & $0.183$ & -- & $0.180$ & --- & $0.245$ & --- & $0.193$ & --- \\
        &Random & $0.336$ & -- & $0.354$ & --- & $0.311$ & --- & $0.336$ & --- \\
        \midrule
        \midrule
        \multirow{6}{*}{\rotatebox[origin=c]{90}{Vanilla}}
        &Llama2-Chat-7B & $0.331_{\pm0.042}$ & 0.0\% & $0.307_{\pm0.024}$ & 0.0\% & $0.229_{\pm0.017}$ & 0.0\% & $0.198_{\pm0.036}$ & 0.0\% \\
        &Llama2-Chat-13B & $0.244_{\pm0.038}$ & 0.0\% & $0.295_{\pm0.019}$ & 0.0\% & $0.213_{\pm0.045}$ & 0.0\% & $0.204_{\pm0.028}$ & 0.0\% \\
        &Llama2-Chat-70B & $0.506_{\pm0.034}$ & 0.0\% & $0.506_{\pm0.044}$ & 0.0\% & $0.368_{\pm0.065}$ & 0.0\% & $0.453_{\pm0.047}$ & 0.0\% \\
        &Mixtral-8x7B & $0.598_{\pm0.050}$ & 0.0\% & $0.509_{\pm0.025}$ & 0.0\% & $0.394_{\pm0.053}$ & 0.0\% & $0.506_{\pm0.059}$ & 0.0\% \\
        &GPT-3.5-Turbo & $0.476_{\pm0.035}$ & 0.0\% & $0.474_{\pm0.028}$ & 0.0\% & $0.262_{\pm0.045}$ & 0.002 & $0.373_{\pm0.020}$ & 0.0\% \\
        &GPT-4-Turbo & $0.682_{\pm0.030}$ & 0.4\% & $0.633_{\pm0.049}$ & 0.0\% & $0.557_{\pm0.036}$ & 0.4\% & $0.666_{\pm0.041}$ & 0.2\% \\
        \midrule
        \midrule
        \multirow{6}{*}{\rotatebox[origin=c]{90}{CoT}}
        &Llama2-Chat-7B & --- & 84.6\% & --- & 79.8\% & --- & 95.4\% & --- & 82.2\% \\
        &Llama2-Chat-13B & $0.367_{\pm0.081}$ & 75.4\% & $0.398_{\pm0.068}$ & 59.6\% & --- & 91.4\% & $0.391_{\pm0.054}$ & 67.0\% \\
        &Llama2-Chat-70B & $0.549_{\pm0.063}$ & 61.8\% & $0.511_{\pm0.058}$ & 66.4\% & --- & 83.2\% & $0.488_{\pm0.053}$ & 73.2\% \\
        &Mixtral-8x7B & $0.670_{\pm0.057}$ & 26.0\% & $0.549_{\pm0.027}$ & 24.0\% & $0.496_{\pm0.067}$ & 21.4\% & $0.543_{\pm0.037}$ & 22.6\% \\
        &GPT-3.5-Turbo & $0.595_{\pm0.032}$ & 0.4\% & $0.503_{\pm0.021}$ & 0.0\% & $0.327_{\pm0.038}$ & 0.4\% & $0.456_{\pm0.050}$ & 0.2\% \\
        &GPT-4-Turbo & $0.883_{\pm0.015}$ & 0.6\% & $0.790_{\pm0.054}$ & 0.0\% & $0.670_{\pm0.044}$ & 0.4\% & $0.823_{\pm0.024}$ & 0.2\% \\
        \midrule
        \midrule
        \multirow{3}{*}{\rotatebox[origin=c]{90}{SimToM}}
        &Mixtral-8x7B & $0.683_{\pm0.055}$ & 10.2\% & $0.617_{\pm0.034}$ & 15.4\% & $0.490_{\pm0.027}$ & 28.2\% & $0.489_{\pm0.045}$ & 18.0\% \\
        &GPT-3.5-Turbo & $0.599_{\pm0.048}$ & 0.0\% & $0.480_{\pm0.024}$ & 0.0\% & $0.248_{\pm0.062}$ & 0.0\% & $0.422_{\pm0.040}$ & 0.0\% \\
        &GPT-4-Turbo & $0.692_{\pm0.039}$ & 0.0\% & $0.743_{\pm0.025}$ & 0.0\% & $0.563_{\pm0.056}$ & 0.0\% & $0.654_{\pm0.028}$ & 0.0\% \\
        \bottomrule
    \end{tabular}}
    \caption{Breakdown of LLMs' performance on the \texttt{MHop} questions. \textit{F1} is the macro F1 score and \textit{Crp.} is the corruption rate. We do not report the F1 score of questions with high corruption rate ($>80\%$).}
    \label{tab:mhop_breakdown}
    \vspace{-0.75em}
\end{table*}

\subsection{Detailed Baseline Results}
The generated responses from LLMs using advanced prompting techniques such as CoT and SimToM are oftentimes in free form. To obtain the final answer, we employed strict parsing rules to extract answer from free-form responses. Any answer that contains ambiguous response or fails to follow the formatting instruction in the prompt are classified as \textbf{corrupted output}. Such results are excluded when computing the accuracy and F1 scores. We provide the \textbf{corruption rate} for each model and prompting method. 

All these details are shown in Table~\ref{tab:detailed_main_results}. For CoT prompting, we do not evaluate $\texttt{Loc}_{f}$ on Llama-Chat models due to their incapability of generating reliable reasoning chains (see corruption rate in Table~\ref{tab:detailed_main_results}). Further, we do not report Llama2-Chat's performance on \texttt{Att} questions due to their high corruption rate. In addition, the SimulatedToM prompting strategy is not evaluated on Llama2-Chat models because of their incompetency in generating character-centric narratives.

Further, as mentioned in \S\ref{sec:question_genres}, we ask two types of questions in \texttt{MHop}, namely questions regarding the \textit{fullness} of a container and questions regarding the \textit{accessibility} of an entity. We show a breakdown of LLMs' performance in each of these sub-tasks in Table~\ref{tab:mhop_breakdown}. We do not report F1 scores for questions with high corruption rate ($>80\%$).

\begin{sidewaystable*}
    \centering
    \begin{tabularx}{0.98\textheight}{P{1em} c | c c c c c c c c c c c c }
    \toprule
        && \multicolumn{12}{c}{\textbf{Large Language Models}} \\[0.5em]
        && \multicolumn{6}{c}{Llama2-Chat} & \multicolumn{2}{c}{Mixtral-Instruct} & \multicolumn{2}{c}{GPT-3.5-Turbo} & \multicolumn{2}{c}{GPT-4-Turbo} \\
    & \# Params & \multicolumn{2}{c}{7B} & \multicolumn{2}{c}{13B} & \multicolumn{2}{c}{70B} & \multicolumn{2}{c}{8x7B} & \multicolumn{2}{c}{---} & \multicolumn{2}{c}{---} \\
    \cmidrule(lr){3-4}  \cmidrule(lr){5-6} \cmidrule(lr){7-8} \cmidrule(lr){9-10} \cmidrule(lr){11-12} \cmidrule(lr){13-14}
    && F1 &  Crp. & F1 & Crp. & F1 & Crp. & F1 & Crp. & F1 & Crp & F1 & Crp. \\
    \midrule
    \multirow{7}{*}{\rotatebox[origin=c]{90}{Vanilla Prompt}} &
        $\texttt{Loc}_{c}$ (F) & $0.290_{\pm0.045}$ & 0.0\% & $0.391_{\pm0.022}$ & 0.0\% & $0.413_{\pm0.016}$ & 0.0\% & $0.512_{\pm0.044}$ & 0.4\% & $0.439_{\pm0.025}$ & 0.0\% & $0.643_{\pm0.061}$ & 0.0\% \\
        & $\texttt{Loc}_{c}$ (S) & $0.462_{\pm0.069}$ & 0.0\% & $0.355_{\pm0.043}$ & 0.0\% & $0.280_{\pm0.028}$ & 0.0\% & $0.294_{\pm0.025}$ & 0.0\% & $0.323_{\pm0.039}$ & 0.0\% & $0.442_{\pm0.044}$ & 0.0\% \\
        & $\texttt{Loc}_{f}$ (F) & $0.404_{\pm0.029}$ & 0.0\% & $0.545_{\pm0.023}$ & 0.0\% & $0.534_{\pm0.023}$ & 0.0\% & $0.399_{\pm0.015}$ & 0.2\% & $0.515_{\pm0.012}$ & 0.3\% & $0.507_{\pm0.010}$ & 0.2\% \\
        & $\texttt{Loc}_{f}$ (S) & $0.245_{\pm0.015}$ & 0.0\% & $0.301_{\pm0.006}$ & 0.0\% & $0.223_{\pm0.023}$ & 0.0\% & $0.211_{\pm0.011}$ & 0.0\% & $0.286_{\pm0.006}$ & 0.0\% & $0.269_{\pm0.004}$ & 0.0\% \\
        & $\texttt{MHop}$ (F) & $0.322_{\pm0.026}$ & 0.0\% & $0.301_{\pm0.023}$ & 0.0\% & $0.501_{\pm0.026}$ & 0.0\% & $0.556_{\pm0.026}$ & 8.8\% & $0.468_{\pm0.029}$ & 0.0\% & $0.658_{\pm0.034}$ & 0.2\% \\
        & $\texttt{MHop}$ (S) & $0.211_{\pm0.024}$ & 0.0\% & $0.229_{\pm0.037}$ & 0.0\% & $0.434_{\pm0.048}$ & 0.0\% & $0.474_{\pm0.025}$ & 5.7\% & $0.334_{\pm0.025}$ & 0.1\% & $0.637_{\pm0.034}$ & 0.3\% \\
        & $\texttt{Att}$ & $0.240_{\pm0.027}$ & 0.0\% & $0.375_{\pm0.031}$ & 0.0\% & $0.415_{\pm0.051}$ & 0.0\% & $0.476_{\pm0.041}$ & 1.6\% & $0.410_{\pm0.021}$ & 0.0\% & $0.544_{\pm0.060}$ & 0.0\% \\
    \midrule
    \multirow{7}{*}{\rotatebox[origin=c]{90}{CoT Prompt}} &
        $\texttt{Loc}_{c}$ (F) & $0.430_{\pm0.045}$ & 54.8\% & $0.414_{\pm0.018}$ & 41.2\% & $0.453_{\pm0.079}$ & 52.0\% & $0.784_{\pm0.070}$ & 5.2\% & $0.587_{\pm0.042}$ & 0.2\% & $0.942_{\pm0.021}$ & 0.6\% \\
        & $\texttt{Loc}_{c}$ (S) & $0.290_{\pm0.030}$ & 58.2\% & $0.287_{\pm0.043}$ & 55.0\% & $0.316_{\pm0.039}$ & 60.8\% & $0.539_{\pm0.060}$ & 8.0\% & $0.457_{\pm0.045}$ & 1.0\% & $0.828_{\pm0.028}$ & 6.0\% \\
        & $\texttt{Loc}_{f}$ (F) & --- & --- & --- & --- & --- & --- & $0.301_{\pm0.015}$ & 0.2\% & $0.469_{\pm0.017}$ & 0.0\% & $0.450_{\pm0.013}$ & 0.0\% \\
        & $\texttt{Loc}_{f}$ (S) & --- & --- & --- & --- & --- & --- & $0.180_{\pm0.010}$ & 0.0\% & $0.240_{\pm0.010}$ & 0.0\% & $0.187_{\pm0.007}$ & 0.0\% \\
        & $\texttt{MHop}$ (F) & $0.374_{\pm0.071}$ & 82.2\% & $0.392_{\pm0.052}$ & 67.5\% & $0.533_{\pm0.049}$ & 64.1\% & $0.610_{\pm0.030}$ & 25.0\% & $0.547_{\pm0.023}$ & 0.2\% & $0.835_{\pm0.027}$ & 0.3\% \\
        & $\texttt{MHop}$ (S) & $0.379_{\pm0.090}$ & 88.8\% & $0.406_{\pm0.061}$ & 79.2\% & $0.527_{\pm0.057}$ & 78.2\% & $0.551_{\pm0.046}$ & 22.0\% & $0.414_{\pm0.026}$ & 0.3\% & $0.755_{\pm0.029}$ & 0.3\% \\
        & $\texttt{Att}$ & --- & 94.8\% & --- & 94.8\% & --- & 99.6\% & $0.519_{\pm0.066}$ & 22.4\% & $0.446_{\pm0.023}$ & 1.6\% & $0.580_{\pm0.034}$ & 4.0\% \\
    \midrule
    \multirow{7}{*}{\rotatebox[origin=c]{90}{SimToM Prompt}} &
        $\texttt{Loc}_{c}$ (F) & --- & --- & --- & --- & --- & --- & $0.414_{\pm0.016}$ & 0.4\% & $0.635_{\pm0.082}$ & 0.0\% & $0.838_{\pm0.024}$ & 2.8\% \\
        & $\texttt{Loc}_{c}$ (S) & --- & --- & --- & --- & --- & --- & $0.290_{\pm0.030}$ & 0.8\% & $0.400_{\pm0.079}$ & 0.0\% & $0.685_{\pm0.037}$ & 2.4\% \\
        & $\texttt{Loc}_{f}$ (F) & --- & --- & --- & --- & --- & --- & $0.352_{\pm0.019}$ & 0.2\% & $0.518_{\pm0.013}$ & 0.0\% & $0.485_{\pm0.011}$ & 0.0\% \\
        & $\texttt{Loc}_{f}$ (S) & --- & --- & --- & --- & --- & --- & $0.206_{\pm0.014}$ & 0.0\% & $0.261_{\pm0.013}$ & 0.0\% & $0.217_{\pm0.023}$ & 0.0\% \\
        & $\texttt{MHop}$ (F) & --- & --- & --- & --- & --- & --- & $0.650_{\pm0.018}$ & 12.8\% & $0.536_{\pm0.023}$ & 0.0\% & $0.720_{\pm0.030}$ & 0.0\% \\
        & $\texttt{MHop}$ (S) & --- & --- & --- & --- & --- & --- & $0.514_{\pm0.018}$ & 0.0\% & $0.350_{\pm0.039}$ & 0.0\% & $0.631_{\pm0.033}$ & 0.0\% \\
        & $\texttt{Att}$ & --- & --- & --- & --- & --- & --- & $0.404_{\pm0.071}$ & 7.2\% & $0.416_{\pm0.031}$ & 0.0\% & $0.488_{\pm0.044}$ & 0.0\% \\
    \bottomrule
    \end{tabularx}
    \caption{Evaluation results in Macro-averaged F1 scores of the \ours~dataset. Location subscripts, $c$ and $f$, represents \textit{coarse} and \textit{fine} respectively. The capital \textit{F} and \textit{S} in the parenthesis represent \textit{first-order ToM} and \textit{second-order ToM}. \textit{Crp.} is the \textit{corruption rate}.}
    \label{tab:detailed_main_results}
\end{sidewaystable*}

\subsection{Effect of Narrative Length}
\label{app:narrative_length}
To study the influence of narrative length on model performance, we conduct a controlled experiment using the \ours-L Narratives. To generate the \ours-L narratives, we fix all other variables, including character names, traits, preference, and only vary the length of the narrative. The \ours-L narratives are on average 2.5 times longer than the original narratives (Table~\ref{tab:data_statistics})

From results shown in Table~\ref{tab:long_narrative}, we see that the length of the narrative has an overall negative impact on LLMs' performance. One clear trend is that the $\texttt{Loc}_{fine}$ questions become harder to answer in long narratives. This is as expected since finding the exact location of an entity becomes more challenging in lengthy narratives.

Further, we see that there are minor improvements in answering $\texttt{MHop}$ questions. This is because that the \textit{Sally-Anne} test has a simple setup (2 characters, 1 entity, and 2 containers). Therefore, expanding the narrative to 500 tokens would force the model or human writer to write more comprehensive descriptions of the characters' actions and thoughts. This would naturally leads the inclusion of more hints that help in answering the \texttt{MHop} questions.

Based on these results, we hypothesize that long stories oftentimes contain narration that are irrelevant to the N-ToM questions, which makes locating fine-grained information (e.g. $\texttt{Loc}_{fine}$) or interpreting character emotion ($\texttt{Att}$) increasingly difficult. Documenting character's mental state of all granularity using symbolic representation such as graph is a potential remedy. Previously, \citet{sclar-etal-2023-minding} proposes to use character-centric graphs to represent each character's mental state and leverage LLMs to reason about character's perception. Such an approach can be studied further and potentially be used in documenting character mental states in long narratives like \ours. \vspace{0.75em} \\

\subsection{\ours~Faithfulness Study}
\paragraph{Detailed Evaluation Results for Faithfulness Study}
\label{app:faithfulness_numbers}
We show the detailed unfaithfulness rate as well as the number of corrupted tuples for each model in Table~\ref{tab:faithfulness_numbers}.

\begin{table} [H]
    \centering
    \resizebox{\columnwidth}{!}{
    \begin{tabular}{P{0.25cm} c|c|c|c}
        \toprule
        & & First-Order & Second Order & Corruption Rate \\
        \midrule
        \multirow{6}{*}{\rotatebox[origin=c]{90}{Separate}} & Llama2-Chat-7B & 0.802 & 0.598 & 0.223 \\
        & Llama2-Chat-13B & 0.098 & 0.166 & 0.220 \\
        & Llama2-Chat-70B & 0.046 & 0.218 & 0.254 \\
        & Mixtral-8x7B & 0.064 & 0.072 & 0.318 \\
        & GPT-3.5-Turbo & 0.054 & 0.000 & 0.000 \\
        & GPT-4-Turbo & 0.100 & 0.200 & 0.000 \\
        \midrule
        \multirow{4}{*}{\rotatebox[origin=c]{90}{Joint}} & Llama2-Chat & --- & --- & $\sim1.00$ \\
        & Mixtral-8x7B & 0.028 & 0.068 & 0.262 \\
        & GPT-3.5-Turbo & 0.026 & 0.128 & 0.111 \\
        & GPT-4-Turbo & 0.030 & 0.112 & 0.164 \\
        \bottomrule
    \end{tabular}}
    \caption{The unfaithfulness rate and the number of corrupted tuples for each model. The unfaithfulness rate of Joint Llama2-Chat models are not reported as all of the Llama2-Chat models fail to follow the prompt in the joint approach.}
    \label{tab:faithfulness_numbers}
\end{table}

\subsection{Addition Experiments on \texttt{Att} Questions}
\label{app:att_questions}
Being mindful of the challenge that the \texttt{Att} questions bring to the LLMs, we conduct additional experiments to further investigate the potential solution and LLMs' mode of error. 

We first examine the Self-Ask prompting method \cite{press-etal-2023-measuring} on \texttt{Att} questions using the same procedure as \S\ref{sec:result}. The results of Self-Ask prompting versus other prompting methods are shown in Table~\ref{tab:self-ask}.

We further compute the recall of LLMs' answers to \texttt{Att} questions. We find that the recalls are low regardless of the choice of LLMs or prompting strategies. We summarise the recall results in Table~\ref{tab:att_recall}.

\begin{table}
    \centering
    \resizebox{\linewidth}{!}{
    \begin{tabular} { c | c c | c c | c c }
        \toprule
        Prompt & \multicolumn{2}{c|}{Mixtral} & \multicolumn{2}{c|}{GPT-3.5-Turbo} & \multicolumn{2}{c}{GPT-4-Turbo} \\
        & F1 & $\Delta$F1 & F1 & $\Delta$F1 & F1 & $\Delta$F1 \\
        \midrule 
        CoT & 0.519 & \tableinc{+0.043} & 0.446 & \tableinc{+0.036} & \boldmath{$0.580$} & \tableinc{+0.036} \\
        \midrule
        SimToM & 0.404 & \tabledec{- 0.072} & 0.416 & \tablesame{+0.006} & \boldmath{$0.488$} & \tabledec{- 0.056} \\
        \midrule
        Self-Ask & 0.529 & \tableinc{+0.053} & 0.458 & \tableinc{+0.048} & \boldmath{$0.617$} & \tableinc{+0.073} \\
        \bottomrule
    \end{tabular}
    }
    \caption{Macro F1 score of \ours~narratives evaluated using only \texttt{Att} questions with advanced prompting methods including CoT, SimToM, and Self-Ask prompt. The numbers on the right are relative \tableinc{performance gain}, \tabledec{performance degradation}, or \tablesame{equal performance} ($\Delta\text{F1} < 0.010$).}
    \label{tab:self-ask}
\end{table}

\begin{table} [t]
    \centering 
    \resizebox{\linewidth}{!}{
    \begin{tabular} { c | c | c | c }
        \toprule
        \multicolumn{3}{l}{{\color{green} \faRobot} : Vanilla Prompt} & 
        \multicolumn{1}{l}{{\color{red} \faLink} : CoT Prompt} \\
        \multicolumn{3}{l}{{\color{yellow} \faPortrait} : SimToM Prompt} &
        \multicolumn{1}{l}{{\color{blue} \faQuestionCircle} : Self-Ask Prompt} \\
        \toprule
        & \multicolumn{3}{c}{Result on \texttt{Neutral} Attitude} \\
        \cmidrule{2-4}
        & Mixtral & GPT-3.5-Turbo & GPT-4-Turbo \\
        \midrule
        {\color{green} \faRobot} & 0.194 & 0.278 & 0.194 \\
        {\color{red} \faLink} & 0.190 & 0.132 & 0.143 \\ 
        {\color{yellow} \faPortrait} & 0.106 & 0.292 & 0.139 \\
        {\color{blue} \faQuestionCircle} & 0.228 & 0.155 & 0.197 \\
        \midrule\midrule 
        & \multicolumn{3}{c}{Result on \texttt{Positive} Attitude} \\
        \cmidrule{2-4}
        & Mixtral & GPT-3.5-Turbo & GPT-4-Turbo \\
        \midrule
        {\color{green} \faRobot} & 0.206 & 0.220 & 0.264 \\
        {\color{red} \faLink} & 0.364 & 0.170 & 0.391 \\ 
        {\color{yellow} \faPortrait} & 0.130 & 0.226 & 0.185 \\
        {\color{blue} \faQuestionCircle} & 0.351 & 0.212 & 0.500 \\
        \midrule\midrule 
        & \multicolumn{3}{c}{Result on \texttt{Negative} Attitude} \\
        \cmidrule{2-4}
        & Mixtral & GPT-3.5-Turbo & GPT-4-Turbo \\
        \midrule
        {\color{green} \faRobot} & 0.927 & 0.821 & 0.952 \\
        {\color{red} \faLink} & 0.819 & 0.838 & 0.936 \\ 
        {\color{yellow} \faPortrait} & 0.833 & 0.226 & 0.905 \\
        {\color{blue} \faQuestionCircle} & 0.797 & 0.747 & 0.972 \\
        \bottomrule
    \end{tabular}
    }
    \caption{Macro-recall of LLMs' answer to the \textit{Neutral} (top) and \textit{Positive} (bottom) \texttt{Att} questions.}
    \label{tab:att_recall}
\end{table}

Through further analysis, we find that the low recall in classifying \textit{Neutral} actions is correlated to the \textit{mover}'s personality. As mentioned in \S\ref{sec:attitude}, the \textit{mover}'s personality is latent with respect to the \textit{observer}'s perception. In addition, we have taken measures to penalize LLMs from using such spurious correlation (see \S\ref{sec:spurious_cue}). Therefore, leveraging such information is doomed to fail. See Table~\ref{tab:neutral_personality} for the proportion of wrongly classified \textit{Neutral} actions that are correlated to the \textit{mover}'s personality.

\section{Examples of \ours~Narratives}
\label{app:opentom_examples}

We provide 6 examples of the \ours~narrative, one for each personality for each length. These examples are shown in the next page.

\newpage

\begin{tcolorbox}[
    title=Example of \ours~Narrative (Considerate \textit{Mover}),
    breakable,
    float*,
    width=\textwidth,
    colback=white,
    colframe=tale!75!black,
    colbacktitle=tale,
    coltitle=black,
    fonttitle=\bfseries
    ]
Genesis and Felix were the best of friends. They both had a deep love for watermelon. The sweet, juicy fruit was their ultimate delight during the hot summer days. Genesis loved the refreshing taste of watermelon, and Felix couldn't resist its vibrant red color. \\
\\
One day, as fate would have it, both Genesis and Felix found themselves in the den. It was there, in the pantry, that they laid their eyes on a massive watermelon. Their mouths watered at the sight. They were overjoyed! \\
\\
But just as quickly as Felix entered the den, he exited, seemingly disinterested in the watermelon. Little did he know that Genesis had a thoughtful plan brewing in her mind. Knowing that they both adored watermelon, Genesis took it upon herself to move the fruit to the kitchen counter. This way, it would be convenient for both Genesis and Felix to grab a slice whenever they desired. \\
\\
And with that, Genesis carefully placed the watermelon on the kitchen counter, satisfied with her kind gesture. The fruit sat there, waiting patiently for the two friends to reunite and relish in the goodness of watermelon once again.
\end{tcolorbox}

\begin{tcolorbox}[
    title=Example of \ours~Narrative (Inconsiderate \textit{Mover}),
    float*,
    width=\textwidth,
    colback=white,
    colframe=tale!75!black,
    colbacktitle=tale,
    coltitle=black,
    breakable,
    fonttitle=\bfseries]
Diego and Amir were both residents of the same apartment complex. They had known each other for quite some time, but they couldn't be more different in their tastes and preferences. One thing that particularly divided them was their opinion on scarves. Diego despised scarves, finding them to be unnecessary and bothersome. On the other hand, Amir adored scarves, always wearing one to complete his outfit. \\
\\
One sunny afternoon, both Diego and Amir happened to stroll into the patio at the same time. As they approached the central basket, their eyes fell upon a colorful scarf lying inside. Diego's face contorted in disdain while Amir's eyes lit up with delight. \\
\\
In that moment, without exchanging any words, Diego swiftly reached into the basket and snatched the scarf. Amir watched curiously as Diego took a few steps towards a nearby donation bin. With a resolute expression, Diego dropped the scarf into the bin, relieving himself of its presence.\\
\\
And just like that, the scarf that once rested in the patio basket had found a new temporary home in the donation bin, waiting to be discovered by someone who would appreciate its warmth and beauty. Diego turned around to leave the patio, completely unaware that his actions had not gone unnoticed by Amir.
\end{tcolorbox}

\begin{tcolorbox}[title=Example of \ours~Narrative (Negativisitc \textit{Mover}),
    float*,
    width=\textwidth,
    colback=white,
    colframe=tale!75!black,
    colbacktitle=tale,
    coltitle=black,
    breakable,
    fonttitle=\bfseries]
Andrew and Richard were two very different individuals. Andrew loved hats, while Richard despised them. It was a peculiar quirk that set them apart. One sunny afternoon, both Andrew and Richard found themselves in the backyard. As they looked around, they couldn't help but notice a hat trapped inside a glass bottle. \\
\\
Curiosity piqued, Andrew decided to explore further. He stayed in the backyard, studying the hat trapped in the bottle. Richard, on the other hand, chose to leave the backyard and head towards the master bedroom. \\
\\
Andrew was a negativistic person. Knowing Richard's disdain for hats, he saw an opportunity to showcase this unique find. With a mischievous grin, Andrew carefully picked up the bottle and moved it to his own room. He imagined his friends and guests admiring the hat as part of his growing collection. Little did he know, Richard had already left the backyard and had no knowledge of Andrew's actions. \\
\\
And just like that, the hat found a new home, hidden away in Andrew's room. The story ends here, leaving us with the anticipation of what might unfold when Richard discovers Andrew's secret.
\end{tcolorbox}

\begin{tcolorbox}[title=Example of \ours-L Narrative (Considerate \textit{Mover}),
    float*,
    width=\textwidth,
    colback=white,
    colframe=tale!75!black,
    colbacktitle=tale,
    coltitle=black,
    breakable,
    fonttitle=\bfseries]
In a quaint corner of their world, Damien and Gabriella shared a residence and, coincidentally, an aversion to a certain leafy green: cabbage. This mutual sentiment did not arise from a spoken agreement or a shared event; rather, it was one of those unspoken truths that hung in the air, visible in their identical expressions of disdain whenever the vegetable made an appearance. \\
\\
It was on a day like any other that they found themselves entering the lounge at different moments. The room, ordinarily a sanctuary adorned with comfort and personal treasures, harbored a curious anomaly. Amidst the shimmering array of jewels and ornate baubles that filled their treasure chest, lay a singular, vibrant cabbage. The vegetable's presence was stark, almost jarring against the backdrop of metallic luster and gilded heirlooms. \\
\\
Without lingering, Gabriella chose to take her leave from the lounge. The room, with its aberrant content, was less appealing with the cabbage's unexpected cameo. She stepped out, allowing the tranquility of the lounge to close behind her, untouched by her transient visit. \\
\\
Damien, on the other hand, was a character often noted for his considerate nature and his penchant for thoughtful deeds. He harbored a peculiar misunderstanding about Gabriella's palate. In his mind, Gabriella was someone who found a certain pleasure in the consumption of cabbage, despite his own feelings of repulsion toward it. Guided by this inaccurate belief, he saw an opportunity for a courteous gesture. \\
\\ 
With measured care, he approached the out-of-place cabbage, nestled incongruously among jewels and trinkets. He lifted it, almost as if he were transporting something of fragility and value, and made his way to the refrigerator. His intentions were clear and simple: to safeguard the cabbage for what he mistakenly perceived as Gabriella's culinary enjoyment. \\
\\
Gabriella, already absent from the scene, was unaware of Damien's actions in the lounge. She did not observe the considerate relocation of the cabbage, did not bear witness to Damiens' silent show of benevolence. \\
\\
Thus, with Damien's small act of kindness, the cabbage found a new home, chilled and preserved within the confines of the refrigerator. The vegetable, once an interloper among treasures, was now nestled amidst cartons and condiments, in a place of practicality rather than display. \\
\\
The story draws to a close with the cabbage's journey complete. There was no more movement for the cabbage, no further interaction. It was now simply a resident of the refrigerator, quietly existing in the chilled environment, its fate to be determined by future culinary choices or eventual disposal. \\
\\
Time alone stood as the silent observer, holding within its steady march the truth about Gabriella's taste. For the moment, however, the cabbage's saga ended, ensconced in the cool shadows behind the refrigerator door, a silent testament to a misjudged preference and an act of unobserved kindness.
\end{tcolorbox}

\begin{tcolorbox}[title=Example of \ours-L Narrative (Inconsiderate \textit{Mover}),
    float*,
    width=\textwidth,
    colback=white,
    colframe=tale!75!black,
    colbacktitle=tale,
    coltitle=black,
    breakable,
    fonttitle=\bfseries]
In a world where personal preferences are as varied as the hues of a rainbow, Abraham found himself at odds with one particular shade: the vibrant orange of melon flesh. His aversion was notorious among his peers. The mere presence of the fruit within his vicinity was enough to set his jaw in a firm line, a silent testament to his profound dislike. \\
\\
Marcos, a colleague who shared Abraham's workspace, held a starkly contrasting view. His affinity for the sweet, succulent fruit was well-known. Where Abraham would avert his gaze from the melon's bright flesh, Marcos would not hesitate to indulge in the pleasure of consuming it, embracing the experience with an appreciative nod. \\
\\
On an unremarkable morning graced by a generous sun, the pair made their entrance into the office. The day commenced like any other, with the mundane tasks of office life beckoning. Yet, amidst the familiarity, something unusual caught their attention. Poised on a table, within a transparent glass bottle, a lone slice of melon lay in wait, its juices glistening, an unwitting siren's call to those who might find it enticing. \\
\\
A frisson seemed to pass through the air as Abraham's gaze landed on the melon. He rose, his movements measured, crossing the distance to the table. With an expression devoid of expression, he reached out and claimed the glass bottle. There was a decisiveness to his actions, a purpose that required no words to be understood. \\
\\
The office, a hive of activity, hardly paused to notice as Abraham exited with the melon in tow. His destination was a small shed outside, a space far removed from the daily bustle. The door swung open with a creak that was quickly silenced as it closed behind him, the melon now sequestered within. \\
\\
Marcos, who happened to witness the silent procession, watched as his colleague carried out the task. His gaze followed Abraham's retreat until he disappeared from sight, leaving a lingering silence in his wake. \\
\\
The glass bottle, now out of sight and out of mind for most, rested in the shadows of the shed. Inside the office, the day resumed its rhythm, as if the fruit had never been there to begin with. Conversations ebbed and flowed, keyboards clicked in a symphony of productivity, and the sun climbed higher in the sky. \\
\\
The fateful morning when Abraham exiled the slice of melon to the confines of the shed would remain a silent chapter in the story of their workplace. It was an event marked not by fanfare or drama but by the simplicity of a task completed, a preference acted upon, and a curious gaze that held no judgment. \\
\\
And there the tale comes to an end, a slice of life captured, a snapshot of two individuals navigating their differences in a shared space. The fate of the melon, now tucked away in the shed, remained a mystery, a subtle reminder of the diverse palette of human inclination and the quiet moments that unfold around them. 
\end{tcolorbox}

\begin{tcolorbox}[title=Example of \ours-L Narrative (Negativisitc \textit{Mover}),
    float*,
    width=\textwidth,
    colback=white,
    colframe=tale!75!black,
    colbacktitle=tale,
    coltitle=black,
    breakable,
    fonttitle=\bfseries]
In the quaint quarters of a shared apartment, there dwelled two roommates, Hadley and Paxton, whose tastes seldom aligned. Among the myriad of their differing opinions, none was as pronounced as their feelings about a particular hat. This hat, a plain and rather nondescript accessory to most, was the crux of an ongoing discord between the two. It was devoid of extravagant features or bold colors, yet it had somehow become the centerpiece of a silent rivalry. \\
\\
Hadley had always harbored a strong distaste for the hat. It was impossible to pinpoint what exactly spurred such loathing for an inanimate object, but its mere presence in the apartment was enough to spark irritation. Conversely, Paxton cherished the hat with an affection that was palpable. To him, the hat was the epitome of elegance and panache, capable of transforming the mundane into something more refined. \\
\\
The hat's usual resting place was atop a shelf in the pantry, among jars of preserves and boxes of tea-- an odd location for a garment, but a neutral territory of sorts. It sat there, quiet and unassuming, as if it had unwittingly become the silent judge of their ongoing quarrel. \\
\\
One unforeseen day, the peculiar fate of cohabitation saw both Hadley and Paxton simultaneously venture into the pantry. As if drawn by some unseen force, their gaze gravitated towards the container on the shelf where the hat lay in wait. The hat, unaware of its divisive nature, continued to exist simply as it was-- a woven construct of fibers and fabric, void of sentiment or the capacity for mockery. \\
\\
Hadley, with a disposition that often leaned towards the oppositional, felt an urgency to act upon the distaste that bubbled to the surface at the sight of the hat. With a decisiveness that seemed almost impulsive, Hadley reached out, fingers grasping the fabric of the hat, and proceeded with a swift motion toward the trash can. Intent on eradicating the hat and the conflict it symbolized, Hadley moved with a resolve that was unyielding. \\
\\
Paxton, meanwhile, stood rooted in place. The movement, the shift in the environment, seemed to unfold in a surreal tableau, challenging the reality of the moment. There was no anticipatory flinch, no audible gasp-- only the starkness of witnessing an action unfold. \\
\\
And so, it came to pass that the hat journeyed from the safety of its perch to the precipice of the garbage receptacle. The air within the confines of the pantry became thick with an unspoken narrative, each roommate enveloped in the stillness of the aftermath. The once silent witness, the hat, now found itself cast in the role of an unwanted protagonist in the midst of a drama it neither asked for nor understood. The roommates, surrounded by the stark walls and the ambient hum of the refrigerator, stood at an impasse. The main event had come and gone, its silent echoes reverberating in the pantry, a room designed for the storage of sustenance now a stage for a silent standoff, unmarred by further development. The hat's fate was left hanging in the balance, the moment frozen in time, as the narrative closed with the weight of unresolved tension, and the memory of the hat's passage towards the bin.
\end{tcolorbox}

\end{document}